\documentclass{article}

\usepackage{microtype}
\usepackage{graphicx}
\usepackage{subcaption}
\usepackage{booktabs} 

\usepackage{hyperref}

\newcommand{\method}{\textsc{Kore}\xspace}

\newcommand{\mme}{\textsc{Kore}$_{\text{MME}}$\xspace}

\newcommand{\ocr}{\textsc{Kore}$_{\text{OCR\raisebox{1ex}{\tiny{VQA}}}}$\xspace}

\newcommand{\matht}{\textsc{Kore}$_{\text{Math\raisebox{1ex}{\tiny{T}}}}$\xspace}

\newcommand{\hall}{\textsc{Kore}$_{\text{Hall\raisebox{1ex}{\tiny{B}}}}$\xspace}

\newcommand{\ie}{\textit{i.e., }}
\newcommand{\eg}{\textit{e.g., }}

\newcommand{\traindata}{\textsc{Kore-74K}\xspace}

\newcommand{\aug}{\textsc{Kore-augmentation}\xspace}
\newcommand{\con}{\textsc{Kore-constraint}\xspace}

\newcommand{\dataset}{\textsc{Evoke}\xspace}

\usepackage{xspace}

\usepackage{multirow}
\usepackage{xcolor}
\usepackage{colortbl}
\usepackage{graphicx} 
\usepackage{caption} 

\usepackage{enumitem}
\usepackage{makecell}   
\definecolor{verylightgray}{gray}{0.95} 

\definecolor{backblue}{RGB}{210, 230, 250}
\definecolor{knowledgeBlue}{RGB}{193, 214, 243}
\definecolor{knowledgePink}{RGB}{255, 196, 203}

\definecolor{knowledgeYellow}{RGB}{255, 203, 56}

\usepackage{tabularray}

\newcommand{\daugshifted}{\raisebox{0.5\depth}{$\uparrow$}}

\usepackage[most]{tcolorbox}

\newtcbox{\bluebox}{on line, box align=base, colback=knowledgeBlue, colframe=white, size=fbox, arc=3pt, 
before upper=\strut, top=-2pt, bottom=-4pt, left=-2pt, right=-2pt, boxrule=0pt, colupper=black}

\newtcbox{\pinkbox}{on line, box align=base, colback=knowledgePink, colframe=white, size=fbox, arc=3pt, 
before upper=\strut, top=-2pt, bottom=-4pt, left=-2pt, right=-2pt, boxrule=0pt, colupper=black}

\newtcbox{\yellowbox}{on line, box align=base, colback=knowledgeYellow, colframe=white, size=fbox, arc=3pt, 
before upper=\strut, top=-2pt, bottom=-4pt, left=-2pt, right=-2pt, boxrule=0pt, colupper=black}

\newcommand{\makeTeaserFigure}{
  \begin{center}
    \centering
    \includegraphics[width=0.93\linewidth]{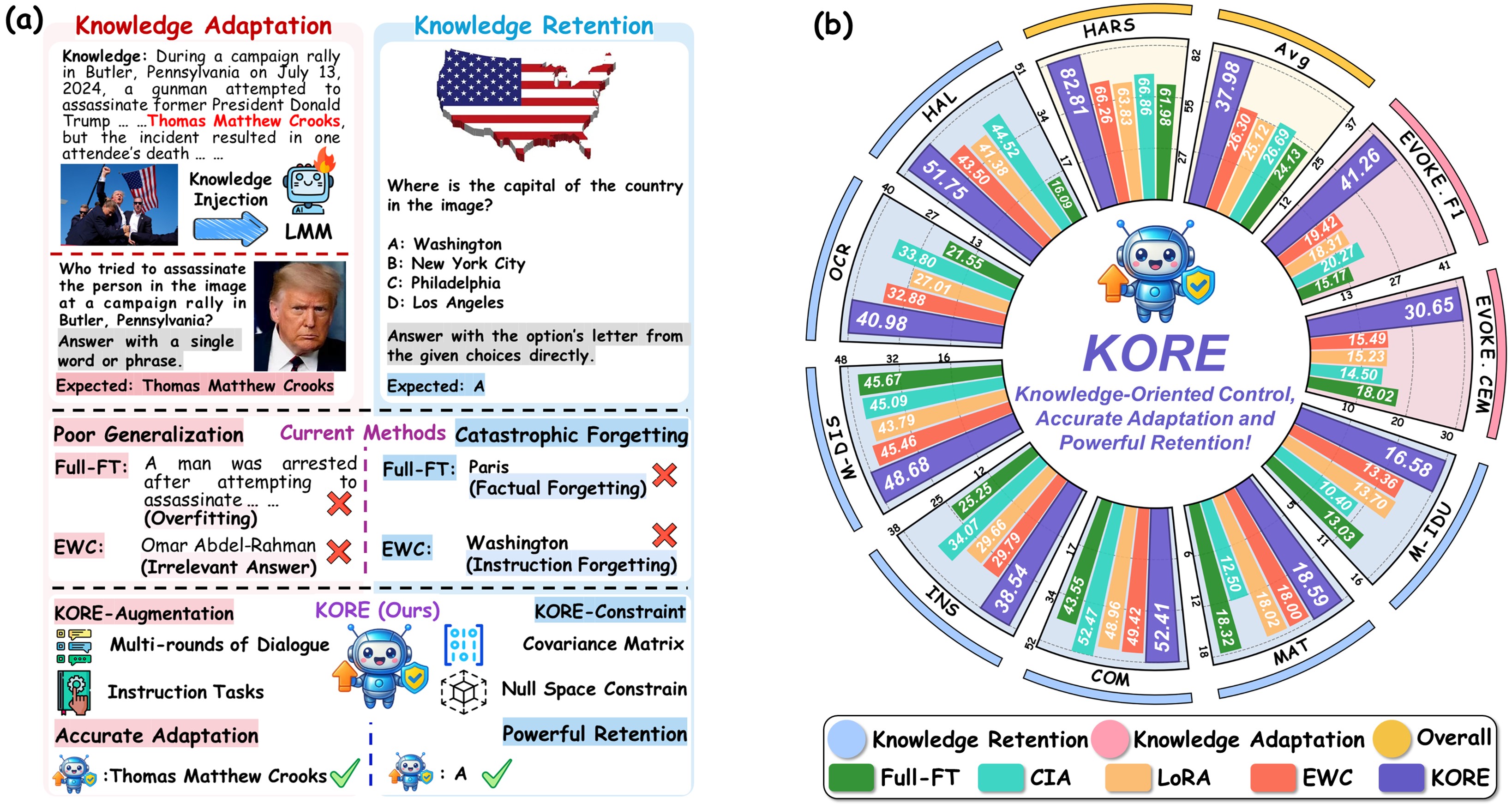}
    
    \captionsetup{hypcap=false}
    \captionof{figure}{\textbf{(a)} Challenges of knowledge adaptation and retention in knowledge injection. While existing methods suffer from poor generalization and catastrophic forgetting, our proposed \method leverages knowledge-oriented controls to achieve both accurate adaptation and powerful retention.
 \textbf{(b)} Performance of various methods on LLaVA-v1.5 {\small(7B)}. \pinkbox{Red},  \bluebox{blue} and \yellowbox{yellow} shading correspond to knowledge adaptation, knowledge retention and overall evaluations, respectively. The definition of overall evaluations' metrics are in \autoref{sec:Experiment}.}
    \label{fig:teaser}
  \end{center}
}

\usepackage{marvosym}
\usepackage{placeins} 

\usepackage{pifont}
\usepackage{adjustbox}

\newcommand{\bfit}[1]{\textbf{\textit{#1}}}

\usepackage[accepted]{icml2026}

\usepackage{amsmath}
\usepackage{amssymb}
\usepackage{mathtools}
\usepackage{amsthm}

\usepackage[capitalize,noabbrev]{cleveref}

\theoremstyle{plain}
\newtheorem{theorem}{Theorem}[section]

\theoremstyle{definition}

\theoremstyle{remark}

\usepackage[textsize=tiny]{todonotes}

\icmltitlerunning{KORE: Enhancing Knowledge Injection for Large Multimodal Models via Knowledge-Oriented Controls}

\begin{document}

\twocolumn[

  \icmltitle{KORE: Enhancing Knowledge Injection for Large Multimodal Models via Knowledge-Oriented Controls}

  \icmlsetsymbol{equal}{*}

  \begin{icmlauthorlist}
    \icmlauthor{Kailin Jiang}{ustc,bigai}
        \icmlauthor{Hongbo Jiang}{xmu}
        \icmlauthor{Ning Jiang}{nefu}
        \icmlauthor{Zhi Gao}{bit,bigai}
        \icmlauthor{Jinhe Bi}{lmu}
        \icmlauthor{Yuchen Ren}{syd}
        \icmlauthor{Bin Li}{ustc}
        \icmlauthor{Yuntao Du}{sdu,nju}
        \icmlauthor{Lei Liu\textsuperscript{\Letter}}{ustc}
        \icmlauthor{Qing Li\textsuperscript{\Letter}}{bigai}
  \end{icmlauthorlist}

  \icmlaffiliation{ustc}{University of Science and Technology of China}
    \icmlaffiliation{bigai}{State Key Laboratory of General Artificial Intelligence, BIGAI}
     \icmlaffiliation{xmu}{Xiamen University}
    \icmlaffiliation{nefu}{Northeast Forestry University}
    \icmlaffiliation{bit}{Beijing Institute of Technology}
    \icmlaffiliation{lmu}{Ludwig Maximilian University of Munich}
        \icmlaffiliation{syd}{The University of Sydney}
    \icmlaffiliation{sdu}{C-FAIR\&school of software, Shandong University}
    \icmlaffiliation{nju}{State Key Lab. for Novel Software Technology, Nanjing University, P.R. China}

    \icmlcorrespondingauthor{Lei Liu}{liulei13@ustc.edu.cn}
    \icmlcorrespondingauthor{Qing Li}{dylan.liqing@gmail.com}

  \icmlkeywords{Machine Learning, ICML}

  \vskip 0.3in

   \makeTeaserFigure
    
   \vskip 0.3in
  
]

\printAffiliationsAndNotice{}  



\begin{abstract}
Large Multimodal Models encode extensive factual knowledge in their pre-trained weights. However, its knowledge remains static and limited, unable to keep pace with real-world developments, which hinders continuous knowledge acquisition. Effective knowledge injection thus becomes critical, involving two goals: knowledge adaptation (injecting new knowledge) and knowledge retention (preserving old knowledge). Existing methods often struggle to learn new knowledge and suffer from catastrophic forgetting. To address these challenges, we propose \textbf{\method}, a synergistic method centered around \textbf{K}n\textbf{O}wledge-o\textbf{R}ient\textbf{E}d controls. These controls are implemented through a two-stage optimization process: (1) \method automatically converts individual knowledge items into structured and comprehensive knowledge to ensure that the model accurately learns new knowledge, enabling accurate adaptation. (2) \method stores previous knowledge in the covariance matrix of LMM's linear layer activations and initializes the adapter by projecting the original weights into the matrix's null space, defining a fine-tuning direction that minimizes interference with previous knowledge, enabling powerful retention. Extensive experiments on various LMMs, including LLaVA-v1.5 {\small(7B)}, LLaVA-v1.5 {\small(13B)}, and Qwen2.5-VL {\small(7B)}, show that \method achieves superior new knowledge injection performance and effectively mitigates catastrophic forgetting.
\end{abstract}


\section{Introduction}

Large Language Models and Large Multimodal Models demonstrate a remarkable ability to store vast world knowledge within their pre-trained weights and recall it during inference \cite{knowledge2,GPT3,knowledge1,liu2024deepseek}. However, their knowledge remains static and fails to keep pace with the evolving real world, leading to outdated responses and an inability to acquire new information continuously. Therefore, effective knowledge injection methods are crucial, enabling models to inject new knowledge while preserving previous knowledge (\eg knowledge adaptation and retention in Figure~\ref{fig:teaser}), thus supporting continuous model evolution \cite{ovadia2024fine,mecklenburg2024injecting}.

The most direct method for injecting new knowledge is full fine-tuning, which updates all model weights. However, this strategy incurs prohibitive computational and storage costs. To address this, Parameter-Efficient Fine-Tuning (PEFT) methods have been introduced for resource-friendly adaptation. PEFT techniques, such as adding adapters \cite{houlsby2019parameter,hu2022lora} or new tokens \cite{lester2021power,sabbatella2024prompt}, drastically reduce the number of trainable parameters by freezing the original pre-trained weights. Despite their success, both full fine-tuning and PEFT methods face significant limitations. They often lead to catastrophic forgetting of pre-existing knowledge and struggle to achieve robust generalization. While full fine-tuning can minimize loss on the training data in \autoref{appendix:loss}, it frequently overfits, failing to effectively extract and manipulate the newly acquired knowledge (\eg Full-FT repeats training data in Figure~\ref{fig:teaser}).

Numerous continual learning techniques, such as rehearsal \cite{li2017learning,hou2019learning} and parameter regularization \cite{kirkpatrick2017EWC,li2017lwf}, have been proposed to mitigate catastrophic forgetting. However, these methods often fail to balance new knowledge acquisition with prior knowledge retention. For example, regularization approaches like EWC \cite{kirkpatrick2017EWC} may impair adaptation to new data, resulting in irrelevant responses and instruction forgetting (\eg EWC leads to irrelevant answer and instruction forgetting in Figure~\ref{fig:teaser}). Drawing inspiration from data augmentation's ability to enhance new knowledge learning \cite{singhal2022large,Allen_Zhu2024PhysicsLM} and continual learning's capacity to preserve old knowledge \cite{MCCLOSKEY1989109,ratcliff1990connectionist}, our proposed \method optimizes the balance between injecting new knowledge and preserving old knowledge, enabling accurate adaptation and powerful retention.

Overall, \method is a synergistic method centered around knowledge-oriented controls, implemented through a two-stage optimization process that integrates \aug and \con. Unlike general augmentation techniques that produce superficial and discrete data variations, \method automatically augments each piece of knowledge into multi-rounds of dialogue and instruction tasks data. This process constructs profound and structured knowledge, which ensures the generalization and internalization of new knowledge and enables the model to flexibly extract and manipulate learned knowledge during inference. Simultaneously, \method stores multimodal knowledge in covariance matrix  $C$ of linear layer activations, assuming $C$ effectively captures previous knowledge (Verification in \autoref{sec:con_analysis}). We then decompose $C$ and extract its null space. Original weights are projected into this null space to initialize a adapter for fine-tuning, which ensures a tuning direction that minimally interferes with the previous knowledge.

To validate the effectiveness of our method, we conducted extensive experiments on multiple representative LMMs. The results in Figure~\ref{fig:teaser} demonstrate that \method exhibits superior performance in both knowledge adaptation and retention compared to standard fine-tuning (\eg Full-FT, LoRA) and continual learning methods (\eg EWC, CIA). Moreover, \method can augment arbitrary knowledge into a structured format and enables customizable knowledge constraints that can be applied based on specific retention needs (\autoref{sec:detailed}). By balancing adaptation and retention through knowledge-oriented controls, \method achieves superior performance without sacrificing flexibility, highlighting its key role in efficient knowledge injection for broader application.

In summary, our main contributions include:

\begin{itemize}[itemsep=7pt, parsep=0pt, topsep=-5pt]
    \item We propose \method, a synergistic method centered on knowledge-oriented controls, addressing the critical trade-off between knowledge internalization and catastrophic forgetting in LMMs during knowledge injection and achieving superior overall performance.

    \item To achieve \textit{accurate adaptation}, we propose \aug, which automatically expands individual knowledge items into multi-round dialogues and instruction-following data, and we further construct the \traindata dataset to serve as a new resource for future knowledge injection research.

\begin{figure*}[t]
  \centering
\includegraphics[width=0.95\linewidth]{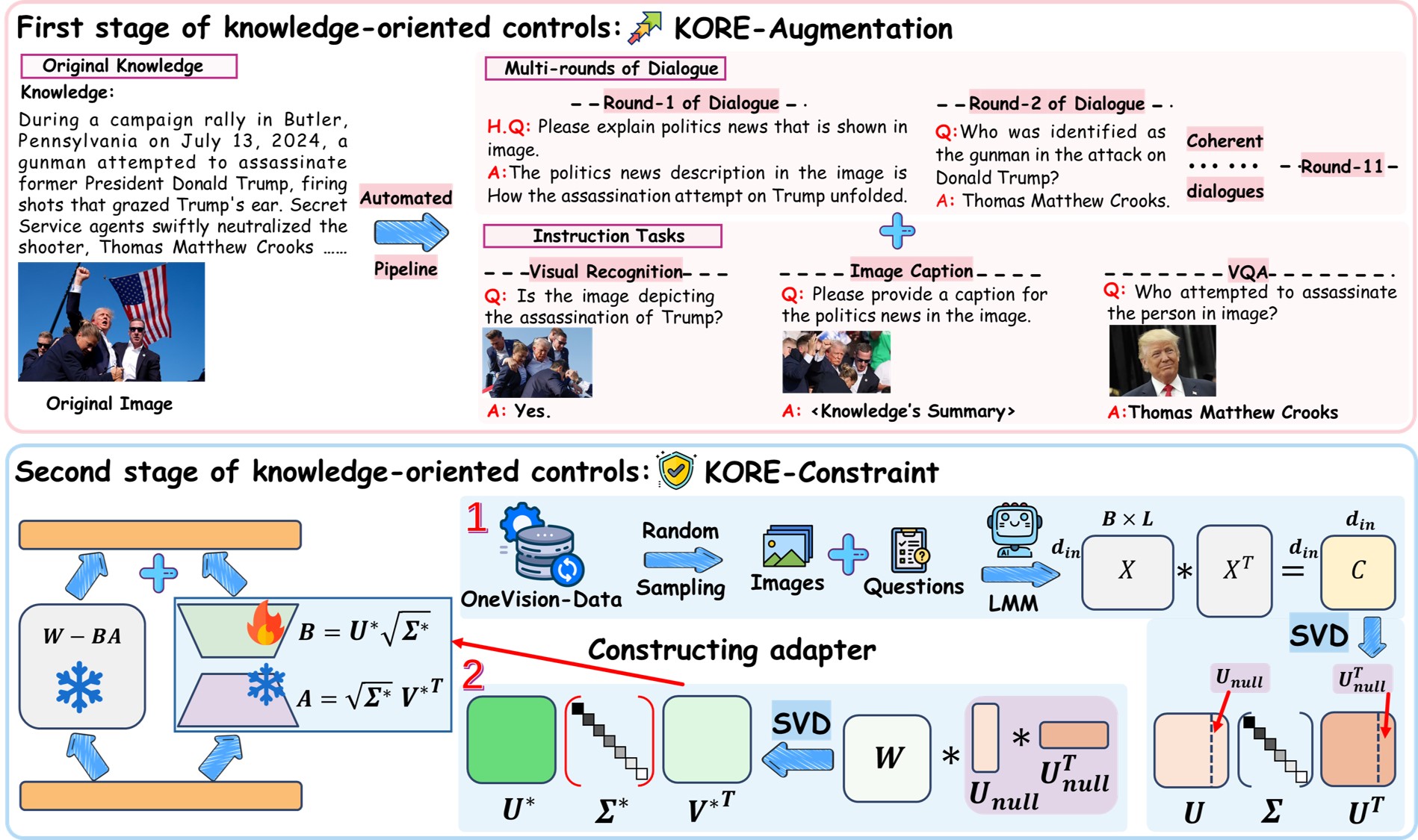}
  \caption{\textbf{Overview.} \method is a synergistic method centered around knowledge-oriented controls, implemented through a two-stage optimization process that integrates \aug and \con. \textbf{(1)} \aug automatically converts each piece of knowledge into profound and structured knowledge. \textbf{(2)} \con minimizes interference with previous knowledge by initializing an adapter with null space that stores covariance matrix of previous knowledge.}
  \label{fig:overview}
\end{figure*}

    \item To ensure \textit{powerful retention}, we introduce \con, which stores previous knowledge in covariance matrix of the model's linear layer activations and initializes the adapter by extracting covariance matrix's null space, thereby defining a fine-tuned direction that minimizes interference with previous knowledge.
\end{itemize}


\section{Related Works}

\subsection{Knowledge Injection}

Injecting new knowledge into models is a critical challenge with two main paradigms. Retrieval-Augmented Generation \cite{song2016better,fan2020enhanced,NEURIPS2020_6b493230}, preserves pre-trained knowledge by leveraging an external knowledge base at inference time, but its efficacy depends on the retrieval system's quality and speed. In contrast, the alternative paradigm directly modifies model parameters, often through efficient methods like full fine-tuning, parameter-efficient fine-tuning \cite{hu2022lora,lauscher2020common}. However, these techniques face a dual challenge, as they often struggle to effectively inject knowledge while still causing catastrophic forgetting \cite{ovadia2024fine,mecklenburg2024injecting}. This highlights a fundamental trade-off between knowledge adaptation and retention, which remains a core problem in knowledge injection.

\subsection{Knowledge Forgetting}

Knowledge injection is fundamentally a continual learning problem that focuses on acquiring new knowledge while retaining prior abilities, ensuring knowledge adaptation without catastrophic forgetting \cite{liu2025continual,huo2025continue,song2025injecting,zheng2025towards}. To tackle these challenges, existing continual learning methods are broadly categorized as follows. Techniques relying on parameter regularization aim to preserve the stability of the model's most critical parameters \cite{kirkpatrick2017EWC,li2017lwf,feng2022overcoming,wang2023orthogonal,liu2024task,qiao2024large,Qi2024InContextEL,qiao2024large,chen2025sefe,liang2025lorasculpt,fang2024alphaedit}. Approaches focused on architecture achieve knowledge retention by introducing either parameter isolation \cite{mallya2018packnet,serra2018overcoming,cao2024generative,zhang2025lori}, adaptive structural elements \cite{yoon2018DEN,hung2019compacting}, or fully modular designs \cite{shen2019progressive}. Rehearsal-based strategies maintain previous capabilities through experience replay utilizing memory buffers \cite{bonicelli2022effectiveness,chen2023dynamic}. Finally, prompt-based methods boost efficiency by employing specific learnable prompts, circumventing the need for explicit data storage \cite{wang2022learning,smith2023coda,wang2022dualprompt}.


\begin{figure*}[t!]
\centering
\includegraphics[width=1\textwidth]{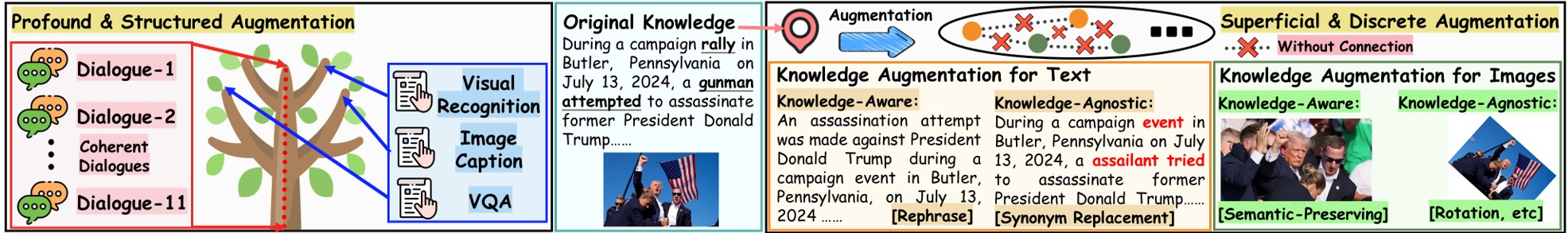}
  \caption{Comparison of \aug \textbf{(left)} and general augmentation methods \textbf{(right)}.}
  \label{fig:augmentation_comparison}
\end{figure*}

\section{Knowledge-Oriented Controls}


\subsection{Knowledge-Oriented Augmentation}
\label{sec:aug}

Existing knowledge injection methods suffer from poor generalization and struggle to master new knowledge \cite{ovadia2024fine,jiang2025when,tang2025evowiki}. Inspired by recent work demonstrating that data augmentation effectively enhances generalization \cite{singhal2022large,Allen_Zhu2024PhysicsLM,wang2025ascdattentionsteerablecontrastivedecoding,park2025new}, we seek to enhance the model's ability to learn new knowledge through data augmentation. However, existing methods are limited to superficial and discrete augmentation, which is insufficient for helping models internalize new knowledge systematically. To address these limitations, we propose \aug, a profound and structured augmentation method via automated pipeline, to build structured and comprehensive knowledge for accurate adaptation.

We observe that \aug augments the original knowledge into multi-rounds dialogues data (forming the trunk) and instruction tasks data (forming the branches), thereby constructing a comprehensive and higher-level knowledge tree (Left part of Figure~\ref{fig:augmentation_comparison}) that supports superior generalization and internalization of new knowledge. \aug moves beyond enabling models to accurately fit training data for ``data memorization''. Instead, it focuses on helping the model comprehend and reason about the inherent logic and associations within the knowledge itself. This enables the model to think, internalize new knowledge, and effectively extract and manipulate the learned knowledge, thereby achieving real ``\textbf{knowledge internalization}''. In contrast, general augmentation methods are superficial and discrete. As shown in right part of Figure~\ref{fig:augmentation_comparison}, for text augmentation, techniques such as knowledge-aware (\eg rephrasing) or knowledge-agnostic (\eg synonym replacement) only create isolated variations. Likewise, image augmentation, whether knowledge-aware (\eg semantic-preserving) or knowledge-agnostic (\eg rotation), operate on a surface level. These methods merely generate isolated data points without connection, superficially modifying existing knowledge to broaden exposure. They fail to construct a coherent knowledge structure. Consequently, general augmentation methods offer limited support for the generalization and internalization of new knowledge. We experimentally validate this statement in \autoref{sec:augmentation} and \aug's implementation details are as follows:


\textbf{Part 1: Constructing Multi-rounds of Dialogue Data.} The multi-rounds of dialogue data for each knowledge sample consists of two components: heuristic Q\&A (H.Q in Figure~\ref{fig:overview}) and dialogue Q\&A. \ding{182} The heuristic Q\&A is constructed randomly using manually written templates. \ding{183} For dialogue Q\&A, we design rigorous rules and diverse task examples, using GPT-4o to generate up to 10  dialogues from original textual knowledge. Ultimately, this process yields 75,710 dialogue data. \textbf{Part 2: Constructing Instruction Tasks Data.} We use News's titles or Entity's names as search key words to retrieve the top five images via Google Search. Visual features of both original and collected images are extracted using CLIP \cite{radford2021learning}. The two images with the highest cosine similarity are retained. \ding{182} \textbf{Visual Recognition:} For this task, questions are randomly selected from a manually written template, and the answer is defined as ``Yes''. One of the previously retained images serves as the query image, accompanied by the instruction, \textit{``Answer this question with Yes or No''}. \ding{183} \textbf{Image Caption:} For this task, answer is a summary generated by GPT-4o based on original textual knowledge. Question is randomly selected from templates, and query image is those remaining from previous steps. And instruction is \textit{``Answer this question in one paragraph''}. \ding{184} \textbf{VQA:} First, we utilize GPT-4o to generate quadruplets $(Q, A, S, H)$ from original textual knowledge, where $Q$ and $A$ form a question-answer pair, $S$ is the subject in question and $H$ is hypernym corresponding to the subject. Subsequently, the subject and hypernym are combined to form a search key words for retrieving and downloading images from Google. The instruction is: \textit{``Answer the question using a single word or phrase''}. This process yields 46,468 VQA samples.

Ultimately, We construct \traindata using original knowledge of EVOKE, and \method is training on \traindata. See more details about \aug in \autoref{appendix:ko_augmentation}.

\begin{figure*}[t!]
    \centering
    \includegraphics[width=1\textwidth]{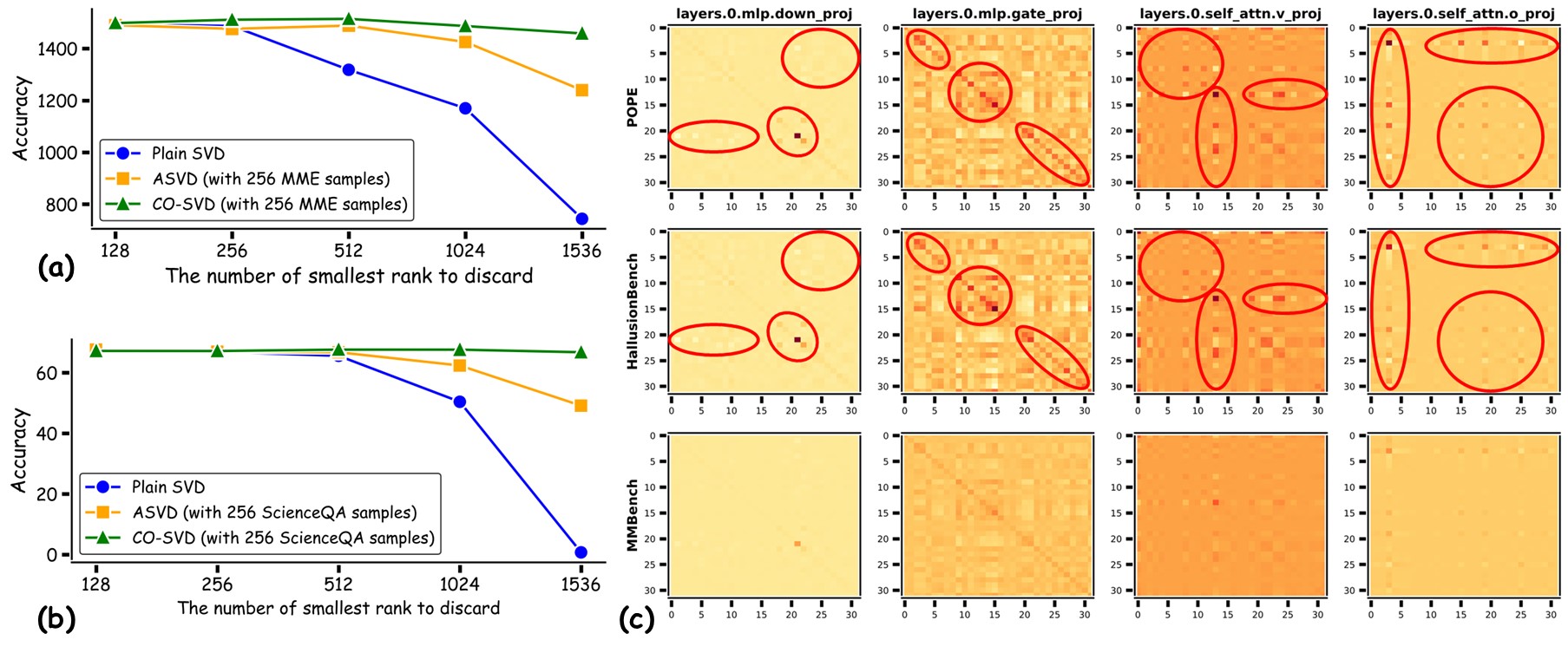}
\vspace{-20pt}
    \caption{Performance (higher is better) on \textbf{(a)} MME and \textbf{(b)} ScienceQA after reconstruction.  \textbf{(c)} Covariance matrix visualization for 4 different input activations in the 0-th block. We down-sample the heatmaps into 32×32. Similar patterns are marked in red circles.}
    \label{fig:capture_know}
\end{figure*}


\subsection{Knowledge-Oriented Constraint}
\label{sec:con}

Large Multimodal Models effectively leverage their pre-trained knowledge to perform a wide range of tasks, and these capabilities are reflected as distinct patterns within their internal activation covariance matrices \cite{MEMIT,yang2024corda}. However, integrating new knowledge or skills into these models presents a fundamental challenge. Direct fine-tuning, the conventional approach, often disrupts these carefully established internal structures, leading to the catastrophic forgetting of prior abilities \citep{rebuffi2017icarl,shi2024continual}. Consequently, the field of continual learning has focused on developing various coSnstraint-based methods to mitigate this performance degradation and preserve foundational knowledge during adaptation.

Inspired by this, we propose \con, a knowledge-oriented constraint method. It stores previous  knowledge in covariance matrix of activations from LMM's linear layers, decomposes this matrix to obtain its null space, and projects the original weights onto this subspace to initialize  adapter. This process ensures that the fine-tuning direction minimally interferes with previous knowledge.

Following prior work \cite{MEMIT,fang2024alphaedit,Tang2025PutTS}, we collect activations from LMMs on a set of random samples representing pre-trained knowledge. Let the input activations to a linear layer be $X \in \mathbb{R}^{d_{in} \times BL}$, where $B$ is the number of samples, $L$ is the sequence length, and $d_{in}$ is the input dimension. And its covariance be $C = XX^T \in \mathbb{R}^{d_{in} \times d_{in}}$. Given pre-trained weights $W_0$, the fine-tuned weights through LoRA are given by: $W^{*} = W_0 + BA$. To achieve knowledge retention, we want to ensure the output activations derived from pretrained knowledge remain consistent after fine-tuning, formalized by the following condition: $W^* C = (W_0 + BA) C \approx W_0 C$. Simplifying this equation further, we obtain: $BA C \approx \mathbf{0}$, and to solve this problem, our goal is to have $A$ located in the null space matrix \cite{Adam-NSCL} of $C$, which is formulated as $A C = \mathbf{0}$. Following the existing methods for conducting null space projection, we first apply a Singular Value Decomposition (SVD) to $C= XX^T$:

\vspace{-10pt}

\begin{equation}
    \text{SVD}\left(X X^T\right) = \sum_{i=1}^{r} \sigma_i \mathbf{u}_i \mathbf{u}_i^T ,
\end{equation}

\vspace{-5pt}

\begin{table*}[t]

  \caption{\textbf{Performance of \method  compared with nine baseline methods.} Row of ``LLaVA-v1.5 {\small(7B)}'' shows performance of pre-trained model.  \textbf{Bold} and \underline{underline} denote the top and runner-up scores, respectively. Results with \colorbox{gray!20}{gray} texture are excluded from sorting.}

  \label{tab:main_result}

    \vspace{-6pt}

  \centering

  \renewcommand{\arraystretch}{1.2} 

  \resizebox{\textwidth}{!}{%

    \begin{tabular}{l|c|cc|ccccccc|cccc}

      \toprule

      \multirow{2.5}{*}{\textbf{Method}}

      & \multirow{2.5}{*}{\textbf{Params}} 

      & \multicolumn{2}{c|}{\textbf{\dataset}}

      & \multirow{2.5}{*}{\textbf{COM} \daugshifted}

      & \multirow{2.5}{*}{\textbf{OCR} \daugshifted}

      & \multirow{2.5}{*}{\textbf{M-DIS} \daugshifted}

      & \multirow{2.5}{*}{\textbf{INS} \daugshifted}

      & \multirow{2.5}{*}{\textbf{M-IDU} \daugshifted}

      & \multirow{2.5}{*}{\textbf{MAT} \daugshifted}

      & \multirow{2.5}{*}{\textbf{HAL} \daugshifted}
         & \multirow{2.5}{*}{\textbf{K.A} \daugshifted}
         & \multirow{2.5}{*}{\textbf{K.R} \daugshifted}

      & \multirow{2.5}{*}{\textbf{Avg} \daugshifted}

       & \multirow{2.5}{*}{\textbf{HARS} \daugshifted} \\

      \cmidrule{3-4} 

      & & \textbf{CEM} \daugshifted & \textbf{F1}\daugshifted  & & & & & & & & \\

      \midrule

      \rowcolor{gray!10}

      LLaVA-v1.5 {\small(7B)} & ---   & 4.89      & 9.34      & 65.61 & 45.59 & 49.22 & 66.33 & 26.37 & 19.33 & 54.32 &7.12&46.74 & 26.93  & --- \\

      \midrule

Full-FT           & 6,759M  & \underline{18.02} & 15.17 & 43.55 & 21.55 & 45.67 & 25.25 & 13.03 & 18.32 & 16.09 &16.60 &31.66 & 24.13  &61.98\\

      LoRA              & 340M    & 15.23 & 18.31 & 48.96 & 27.01 & 43.79 & 29.66 & 13.70 & 18.02 & 41.38 &16.77&33.47 & 25.12  &63.83 \\

      \midrule

      Replay

      & 340M    & 11.36 & 17.98 & 59.72 & 37.98 & \underline{48.64} & \textbf{62.33} & \textbf{19.31} & 19.17 & \underline{51.67} &14.67&\textbf{43.00} &  \underline{28.83}  &66.04 \\

      EWC               & 340M    & 15.49 & 19.42 & 49.42 & 32.88 & 45.46 & 29.79 & 13.36 & 18.00 & 43.50 &17.46&35.14 & 26.30   & 66.26\\

      LwF               & 340M    & 14.58 & 19.99 & 53.14 & 28.77 & 43.41 & 36.19 & 13.68 & 18.22 & 44.18 &17.29&35.44 &  26.36  & 66.26 \\

      MoELoRA           & 340M    & 6.45  & 12.20 & \underline{60.79} & 38.79 & 48.27 & 35.03 & \underline{17.85} & \underline{19.79} & 49.99 &9.33&40.51 & 24.92  & 37.22\\

      O-LoRA            & 340M    & 6.44  & 12.08 & \textbf{61.47}  & \underline{40.91} & 48.07 & 34.85 & 17.28 & \textbf{19.87} &  51.12 &9.26&\underline{41.25} & 25.26  & 36.70\\

      {SEFE}              & 340M    & 16.22 & \underline{24.12} & 36.77 & 35.94 & 48.55 & \underline{42.90} & 10.55 & 18.59 & 31.55 &\underline{20.17}&33.02 & 26.60  & \underline{67.56}\\

      {CIA}       & 340M    & 14.50 & 20.27 & 52.47 & 33.80 & 45.09 & 34.07 & 10.40 & 12.50 & 44.52 &17.39&35.99 & 26.69  & 66.86\\

      \midrule

      \rowcolor{backblue!80}

      \textbf{\method(rank=235)}           & 340M & \textbf{30.65} & \textbf{41.26} & 52.41    &  \textbf{40.98}    & \textbf{48.68}  & 38.54  & 16.58    & 18.59    & \textbf{51.75} &\textbf{35.96}&40.00 & \textbf{37.98}  & \textbf{82.81}\\

        \rowcolor{gray!10}
\textbf{\method(rank=256)}          & 369M &31.05& 41.32&52.48&39.96&48.96&60.02&23.18&18.09&51.50 &36.19&42.10  &39.14  & 84.93\\

      \bottomrule
    \end{tabular}%
  }
\end{table*}

where $U$ is the orthogonal matrix of left singular vectors, respectively, and $\Lambda$ is a diagonal matrix with singular values $\sigma_1 \geq \sigma_2 \geq \dots \geq \sigma_R > 0$ (with $R = \mathrm{rank}(C)$). The null space of $C$ is spanned by $U_{\text{null}} \in \mathbb{R}^{d_{\text{in}} \times (d_{\text{in}} - R)}$, a submatrix containing the last $(d_{\text{in}} - R)$ columns of $U$ that correspond to zero singular values. As shown in the first step on the right side of Figure~\ref{fig:overview}, $U_{\text{null}}$ satisfies $U_{\text{null}}^T C = \mathbf{0}$.

We approximate the null space with $\hat{U} \in \mathbb{R}^{d_{\text{in}} \times r}$, a submatrix containing the $r$ left singular vectors from $U$ associated with the smallest singular values, where $r$ is the predefined LoRA's rank. From this, we define a knowledge-oriented constraint projector $P = \hat{U}\hat{U}^T$. As shown in Figure~\ref{fig:overview}, we then initialize the LoRA adapters by factorizing the pre-trained weights projected into this null space. We compute the SVD of the projected weights: $\text{SVD}\left(W_0 P\right) = \left\{ U^*, \Sigma^*, (V^*)^T \right\}$ and initialize the adapter matrices $B$ and $A$ as:

\vspace{-10pt}

\begin{equation}
    B = U^* \sqrt{\Sigma^*}, \qquad A = \sqrt{\Sigma^*} {(V^*)}^T,
\end{equation}

\vspace{-5pt}

where $\sqrt{\Sigma^*}$ denotes the diagonal matrix with entries for singular values. Finally, to ensure the model is unchanged at the start of fine-tuning, the original weight matrix is adjusted with a residual term:

\vspace{-10pt}

\begin{equation}
W'_0 = W_0 - BA.
\end{equation}

\vspace{-5pt}

Given the asymmetry between $A$ and $B$, fine-tuning only $B$ suffices for strong performance \citep{lora_fa,zhu2024asymmetry}. By freezing $A$ within the null space of $C$, \method ensures $AC \approx \mathbf{0}$, making the update term $BAC$ negligible regardless of any changes to $B$. Proof is in \autoref{appendix:proof}.



\subsection{Analysis of Knowledge-Oriented Constraint}
\label{sec:con_analysis}

\con relies on the premise that the extracted covariance matrix effectively captures knowledge from previous data. Therefore, we expand CO-SVD \cite{yang2024corda} from pure text scenarios to multimodal scenarios to verify \textit{``whether covariance matrices can capture multimodal knowledge and activate distinct modes?''} We apply Plain SVD, ASVD \cite{yuan2023asvd} and CO-SVD to fully decompose all layers of LLaVA-v1.5 {\small(7B)} pre-trained weights. The weights are reconstructed by removing the components corresponding to the $r$ smallest singular values.

Our analysis reveals two key findings: \ding{182} Figure~\ref{fig:capture_know} (a) and (b) demonstrate that CO-SVD exhibits superior performance retention compared to Plain SVD, ASVD and suggest that multimodal knowledge can be effectively captured and stored in covariance matrix. \ding{183} Figure~\ref{fig:capture_know} (c) shows that covariance matrices of linear layer inputs share similar outlier patterns for related tasks (\eg POPE and HallusionBench), but differ from unrelated ones (\eg MMBench), indicating that distinct tasks exhibit different outlier distributions in the covariance matrix. To build a multi-dimensional covariance matrix for \method, we finally sample 256 examples from OneVision \cite{li2025llava-onevision} (General, Doc/Chart/Screen, Math/Reasoning, General OCR). See more results in \autoref{appendix:evoke_presentation}.

\section{\label{sec:Experiment}Experiments}

\begin{table*}[t]
\caption{Performance comparison between \method and baseline methods on fine-grained knowledge retention evaluations. MM\raisebox{1ex}{\tiny{B}}: MMBench; SEED\raisebox{1ex}{\tiny{B2P}}: SEEDBench2\_Plus;  {Math}\raisebox{1ex}{\tiny{T}}: MathVista ; {Math}\raisebox{1ex}{\tiny{I}}: MathVision; {Hall}\raisebox{1ex}{\tiny{B}}: HallusionBench. MME's score is normalized.}
\label{table:main_each_benchmark}
 \vspace{-6pt}

\centering
\resizebox{1\linewidth}{!}{
\begin{tabular}{l|c c|c c|c c|c|c|c c|c c|c}
\toprule
\multirow{2.5}{*}{\textbf{Method}} & \multicolumn{2}{c|}{\textbf{COM}} & \multicolumn{2}{c|}{\textbf{OCR}} & \multicolumn{2}{c|}{\textbf{M-DIS}} & \multicolumn{1}{c|}{\textbf{INS}} & \multicolumn{1}{c|}{\textbf{M-IDU}} & \multicolumn{2}{c|}{\textbf{MAT}} & \multicolumn{2}{c|}{\textbf{HAL}} & \multirow{2.5}{*}{\textbf{Avg}} \\
\cmidrule{2-13}
& \textbf{MME} \daugshifted & \textbf{MM}\raisebox{1ex}{\tiny\textbf{B}} \daugshifted & \textbf{SEED}\raisebox{1ex}{\tiny\textbf{B2P}} \daugshifted & \textbf{OCR}\raisebox{1ex}{\tiny\textbf{VQA}} \daugshifted & \textbf{SQA} \daugshifted & \textbf{MMMU} \daugshifted & \textbf{MIA}\raisebox{1ex}{\tiny\textbf{B}} \daugshifted & \textbf{MMDU} \daugshifted & \textbf{Math}\raisebox{1ex}{\tiny\textbf{T}} \daugshifted & \textbf{Math}\raisebox{1ex}{\tiny\textbf{I}} \daugshifted & \textbf{POPE} \daugshifted & \textbf{Hall}\raisebox{1ex}{\tiny\textbf{B}} \daugshifted \\
\midrule
\rowcolor{gray!10}
LLaVA-v1.5 {\small(7B)} 
& 66.63 & 64.60 & 38.78 & 52.41 & 69.83 & 28.60 & 66.33 & 26.37 & 25.50 & 13.16 & 86.87 & 21.76 & 46.74 \\
\midrule
Full-FT
& 34.17 & 52.92 & 31.44 & 11.65 & 67.13 & 24.20 & 25.25 & 13.03 & 24.70 & 11.94 & 74.22 & 9.27 & 31.66 \\

LoRA
& 44.06 & 53.87 & 30.22 & 23.80 & 66.18 & 21.40 & 29.66 & 13.70 & 23.20 & 12.83 & 73.97 & 8.78 & 33.47 \\

\midrule

Replay
& \underline{58.96} & 60.48 & \underline{38.34} & 37.73 & 68.77 & 28.50 & \textbf{62.33} & \textbf{19.31} & 25.20 & \underline{13.13} & \textbf{85.44} & 17.90 & \textbf{43.00} \\

EWC
& 48.57 & 50.26 & 33.60 & 32.16 & 65.71 & 25.20 & 29.79 & 13.36 & 23.30 & 12.76 & 76.22 & 10.77 & 35.14 \\

LwF
& 50.87 & 55.41 & 32.02 & 25.52 & 66.21 & 20.60 & 36.19 & 13.68 & 24.40 & 12.04 & 79.23 & 9.13 & 35.44 \\

MoELoRA
& 58.26 & \textbf{63.32} & 37.42 & 40.17 & \textbf{69.04} & 27.50 & 35.03 & \underline{17.85} & \underline{27.80} & 11.78 & 80.70 & 19.29 & 40.51 \\

O-LoRA 
& \textbf{60.30} & \underline{62.63} & 37.90 & \underline{43.91} & \underline{68.84} & 27.30 & 34.85 & 17.28 & \textbf{28.20} & 11.55 & \underline{81.46} & \underline{20.78} & \underline{41.25} \\

SEFE 
& 17.27 & 56.27 & \textbf{39.00} & 32.88 & 66.10 & \textbf{31.00} & \underline{42.90} & 10.55 & 23.90 & \textbf{13.29} & 50.61 & 12.48 & 33.02 \\

CIA  
& 50.65 & 54.30 & 33.29 & 34.31 & 67.28 & 22.90 & 34.07 & 10.40 & 24.00 & 11.60 & 79.29 & 9.75 & 35.99 \\

\midrule
\rowcolor{backblue!80}
\textbf{\method(rank=235)}  
& 49.84 & 54.98 & 37.73 & \textbf{44.24} & 68.06 & \underline{29.30} & 38.54 & 16.58 & 25.10 & 12.09 & 80.99 & \textbf{22.51} & 40.00 \\

        \rowcolor{gray!10}
\textbf{\method(rank=256)}  &50.06&54.90&36.89&43.03&68.51&29.40&60.02&23.18&24.70&11.48&80.77&22.23&42.10 \\

\bottomrule
\end{tabular}
}

\end{table*}

\begin{figure*}[t!]
\centering
\includegraphics[width=0.95\textwidth]{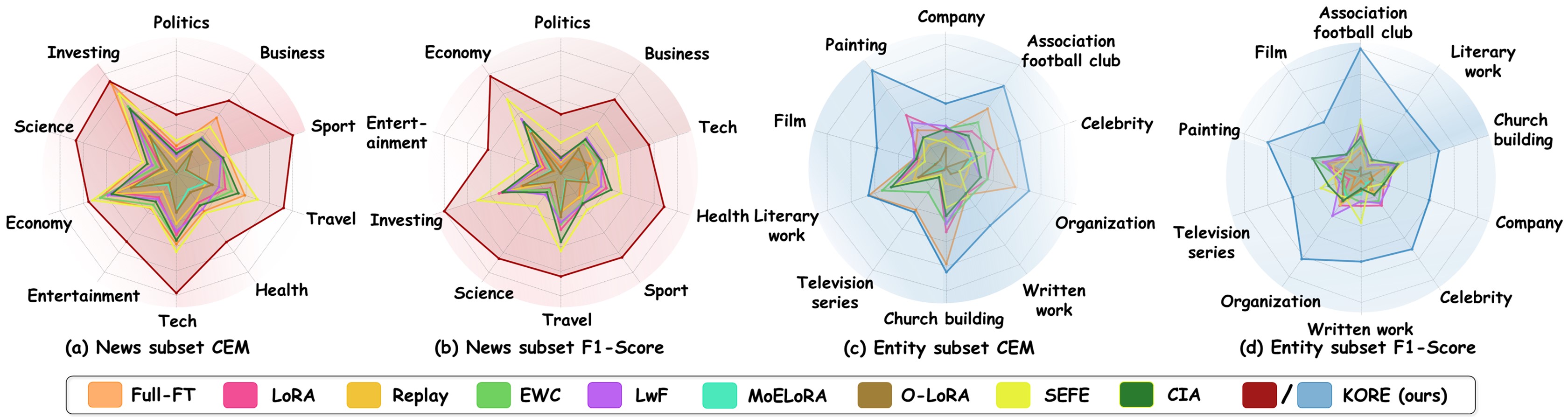}
\vspace{-5pt}
  \caption{{Comparison between \method and baseline methods on fine-grained knowledge types.}
}
  \label{fig:know_types}
\end{figure*}

In this section, we introduce experimental content. See more details and evaluation protocol in \autoref{appendix:setup}.


\textbf{Setup 1: Knowledge Adaptation Evaluation.} we evaluate LMMs by fine-tuning them on EVOKE \cite{jiang2025when}, where knowledge is injected as an image-text pair, with evaluation questions derived from the text. 


\textbf{Setup 2: Knowledge Retention Evaluation.} We evaluate fine-tuned LMMs on 12 benchmarks across 7 capability dimensions, covering the following tasks: \ding{182} {Comprehensive Evaluation (COM):} {MME} \cite{fu2023mme} and {MMBench} \cite{liu2023mmbench}; \ding{183} {Optical Character Recognition (OCR):} {SEEDBench2\_Plus} \cite{li2024seed2plus} and {OCRVQA} \cite{mishra2019ocr}; \ding{184} {Multidisciplinary Reasoning (M-DIS):} {ScienceQA} \cite{lu2022scienceqa} and {MMMU} \cite{mmmu}; \ding{185} {Instruction Following (INS):} {MIA-Bench} \cite{qian2024mia}; \ding{186} {Multi-Turn Multi-Image Dialog Understanding (M-IDU):} {MMDU} \cite{mmdu}; \ding{187} {Mathematical Reasoning (MAT):} {MathVista} \cite{mathvista} and {MathVision} \cite{math_vision}; \ding{188} {Hallucination (HAL):} {POPE} \cite{pope} and {HallusionBench} \cite{hallusionbench}. 

\textbf{Setup 3: Metrics for Overall Evaluations.} \ding{182} \textit{\textbf{Avg}} score is the average performance of knowledge adaptation and retention (\ie K.A and K.R). \ding{183} \textit{\textbf{HARS}} denotes the Harmonized Adaptation-Retention Score, which draws an analogy to the balance between Precision and Recall by integrating knowledge adaptation and retention into a unified metric to evaluate their equilibrium through the following formula:

\vspace{-10pt}
\begin{equation}
    HARS = 2 \cdot \frac{f_A \cdot f_R}{f_A + f_R}
\end{equation}

where $f_A = \frac{G_A}{G_A + 100} \times 100$ and $f_R = G_R + 100$ denote the normalized scores for adaptation and retention, respectively. These scores are derived from the relative gains $G_A = (K.A - K.A_0)/K.A_0$ and $G_R = (K.R - K.R_0)/K.R_0$, where $K.A_0$ and $K.R_0$ denote knowledge adaptation and retention performance of the pre-trained model.


\textbf{Setup 4: Baseline Methods.} We compare against several baselines: Full-FT, LoRA, Replay, EWC, LwF \cite{li2017lwf}, MoELoRA \cite{MoELoRA2024}, O-LoRA \cite{wang2023orthogonal}, SEFE \cite{chen2025sefe} and CIA \cite{qiao2024large}.



\subsection{Analysis of main results}\label{sec:main}

We provide the performance of various methods in Table~\ref{tab:main_result} and more results, case studies in \autoref{appendix:results} and \autoref{appendix:case}, from which we draw the following observations: \ding{182} \method enables accurate adaptation for effectively injecting new knowledge. Specifically, \method (rank=235) achieves improvements of $12.63\uparrow$ in CEM and $17.14\uparrow$ in F1-Score over the best baseline on \dataset, even outperforming LoRA by more than twofold in K.A. \ding{183} \method enables powerful retention for effectively preserving previous knowledge. Specifically, \method (rank=235) outperforms Full-FT, LoRA, EWC, LwF, SEFE, CIA in K.R and achieves top scores on OCR, M-DIS, HAL. \ding{184} \method achieves remarkable holistic performance by harmonizing the dual objectives of knowledge injection. Specifically, \method (rank=235) achieves improvements of $9.15\uparrow$ in Avg and $15.25\uparrow$ in HARS over the best baseline, demonstrating its superior overall performance.


\subsection{Analysis of fine-grained results}\label{sec:detailed}


According to the fine-grained knowledge adaptation and retention performance shown in Table~\ref{table:main_each_benchmark} and Figure~\ref{fig:know_types}, we draw the following observations: \ding{185} \method achieves competitive knowledge retention. Specifically, \method outperforms LoRA (\eg $6.53\uparrow$ in Avg) and continual learning methods (\eg EWC, LwF, SEFE and CIA), achieving superior scores on OCR\raisebox{1ex}{\tiny{VQA}}, MMMU and Hall\raisebox{1ex}{\tiny{B}}. Furthermore, by adjusting trainable parameters (rank=256) and covariance matrix source (Table~\ref{table:specific_each_benchmark}), it closely matches or even exceeds Replay. \ding{186} \method demonstrates superior performance across a wide spectrum of fine-grained knowledge. Figure~\ref{fig:know_types} compares 20 fine-grained News and Entity types from \dataset. \method consistently outperforms all baselines, demonstrating strong and comprehensive knowledge adaptation.

 \vspace{-10pt}

\begin{figure}[h!] 
  \centering
  \includegraphics[width=0.95\linewidth]{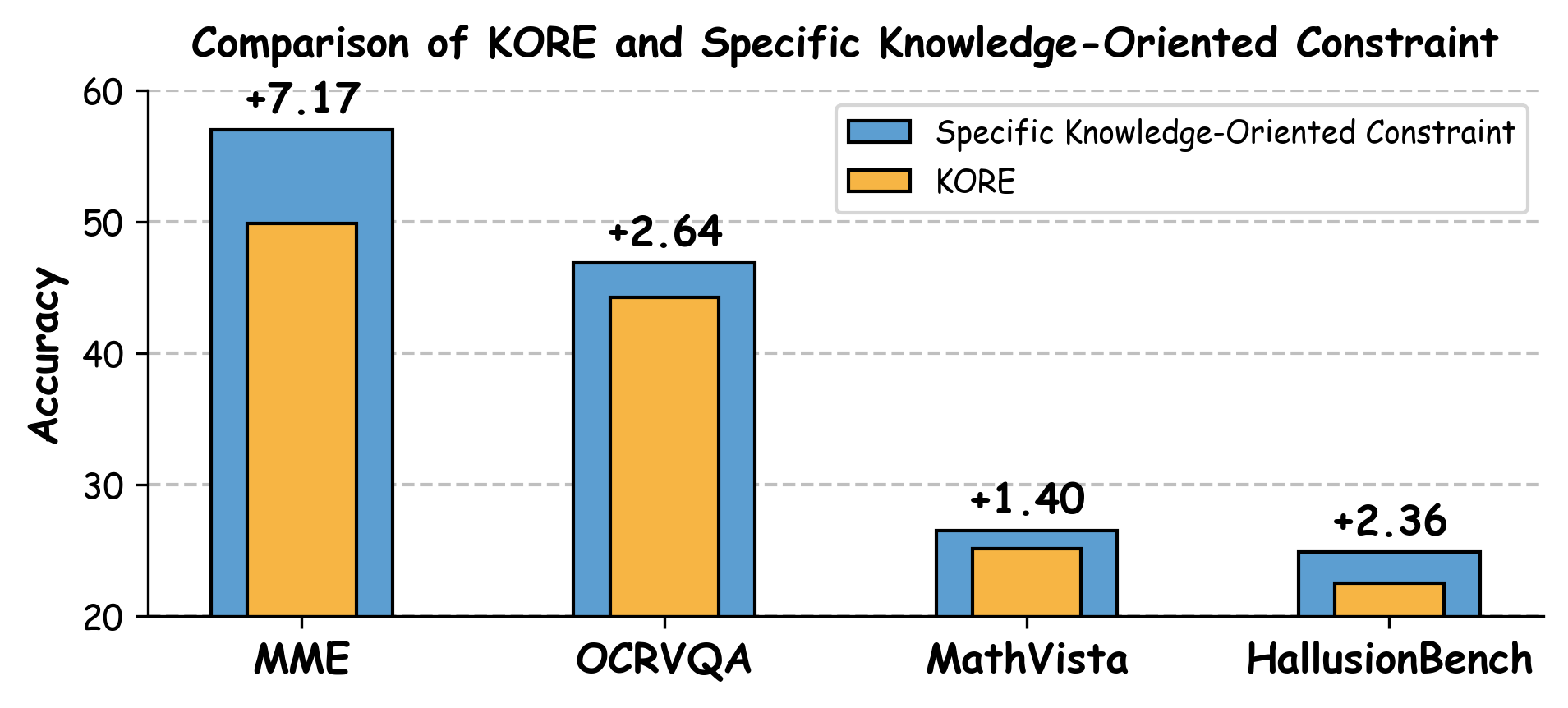}
  \vspace{-8pt}
    \caption{Performance comparison of corresponding tasks under specific knowledge-oriented constraints.}
    \label{fig:specific_constrain}
\end{figure}

\begin{figure*}[t!] 
  \centering
  \includegraphics[width=0.95\linewidth]{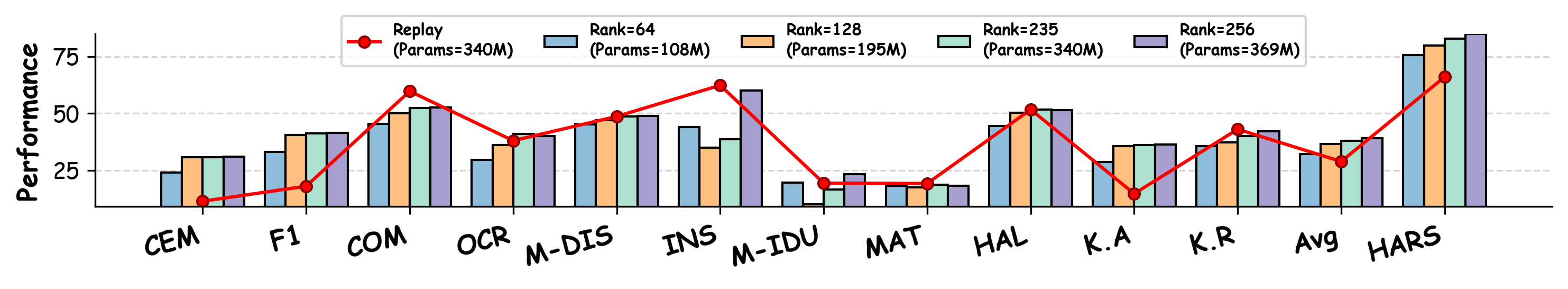}
  \vspace{-8pt}
  \caption{Comparison of different ranks for \method with LLaVA-v1.5 {\small(7B)}.}
  \label{fig:rank_final}
\end{figure*}

 \vspace{-8pt}

Given the diverse prior knowledge of LMMs, we investigate  \textbf{\textit{whether \method can preserve specific knowledge without compromising new knowledge injection or other existing abilities?}} We construct specific constraints by sampling 256 data per benchmark across four dimensions. \ding{187} Specific constraints enhance specific knowledge retention. Figure~\ref{fig:specific_constrain} shows that specific constraints enhance targeted knowledge retention, notably with a $7.17\uparrow$ gain on MME, demonstrating their potential for tailored knowledge retention.


\begin{table}[t!]
    \centering
  \caption{Performance comparison between \method and baseline methods with various LMMs scales and architectures.}
  \vspace{-6pt}
  \label{tab:scales_architectures}
    \renewcommand{\arraystretch}{1.2}
    \resizebox{0.95\linewidth}{!}{%
        \begin{tabular}{l|c|c|c|c}
            \toprule
            \textbf{Method} & \textbf{K.A} \daugshifted & \textbf{K.R} \daugshifted & \textbf{Avg} \daugshifted & \textbf{HARS} \daugshifted \\

            \midrule

     
      \rowcolor{gray!10}

LLaVA-v1.5 {\small(13B)}  &7.70&49.51&28.60& --- \\ 
LoRA  &\underline{19.55}&32.64&26.09&63.15 \\  

 Replay  &16.13&\textbf{45.69}&\underline{30.91}&\underline{66.73} \\   
\rowcolor{backblue!80}
      \textbf{\method} &\textbf{38.68}&\underline{45.35}&\textbf{42.01}&\textbf{85.46} \\  

      \midrule
      \rowcolor{gray!10}

Qwen2.5-VL {\small(7B)}  &12.34&66.89&39.61& ---\\ 
LoRA  &14.29&38.16&26.22&22.03 \\  

 Replay  &\underline{15.12}&\textbf{63.94}&\underline{39.53}&\underline{30.89} \\   
\rowcolor{backblue!80}
      \textbf{\method} &\textbf{27.14}&\underline{58.31}&\textbf{42.72}&\textbf{67.10} \\

            \bottomrule
        \end{tabular}%
    }
\end{table}

\begin{table}[t!]
    \centering
    \caption{Comparison of ablation experiment results of \method.}
    \vspace{-6pt}
    \label{tab:ablation_main}
    \renewcommand{\arraystretch}{1.2}
    \resizebox{0.95\linewidth}{!}{%
        \begin{tabular}{l|c|c|c|c}
            \toprule
            \textbf{Method} & \textbf{K.A} \daugshifted & \textbf{K.R} \daugshifted & \textbf{Avg} \daugshifted & \textbf{HARS} \daugshifted \\  
            \midrule

            \rowcolor{backblue!80}
            \textbf{\method} & 35.96 & \underline{40.00} & \textbf{37.98} & \textbf{82.81} \\  
            W/o Augmentation & 14.57 & \textbf{40.16} & 27.37 & 64.14 \\ 
            W/o Constraint   & \textbf{38.82} & 35.78 & 37.30 & 79.04 \\ 
            W/o Frozen Matrix $A$ & \underline{36.85} & 38.92 & \underline{37.88} & \underline{81.96} \\ 

            \bottomrule
        \end{tabular}%
    }
\end{table}



\subsection{Analysis of various LMM scales and architectures}\label{sec:scales}

We evaluate the universality and robustness of \method on larger and architecturally distinct models, using Replay and LoRA as baseline methods. And we draw the following observations: \ding{188} \method shows enhanced superiority on a larger-scale LMM. Table~\ref{tab:scales_architectures} shows that \method surpasses LoRA across all dimensions, achieving improvements of $11.10\uparrow$ in Avg and $18.73\uparrow$ in HARS over Replay. \ding{189} \method’s effectiveness is not architecture-specific. On Qwen2.5-VL {\small(7B)}, \method achieves improvements of $3.19\uparrow$ in Avg and $36.21\uparrow$ in HARS  over Replay (proving the necessity of HARS metric), demonstrating that it consistently delivers superior performance across various architectures.


\subsection{Analysis of ablation experiments}\label{sec:ablation}

In this section, we conduct extensive ablation studies (\eg Rank, W/o Augmentation, W/o Constraint and W/o Frozen Matrix $A$) to validate the effectiveness of \method's design. 

\ding{190} Larger rank enhance \method's performance. Figure~\ref{fig:rank_final} shows a clear trend: \method's performance increases with higher rank and more trainable parameters on nearly all evaluations. \method (rank=64) still surpasses Replay in K.A, Avg and HARS, only using less than half of parameters of Replay. \ding{191} Ablation studies reveals the effectiveness of \method's design. Table~\ref{tab:ablation_main} validates the design of \method, demonstrating that the two-stage optimization design centered on knowledge-oriented controls contributes positively to overall performance. W/o Augmentation is particularly detrimental to knowledge adaptation ($21.39 \downarrow$ in K.A). Meanwhile, W/o Constraint and W/o Frozen Matrix $A$ impairs knowledge retention.

\vspace{-2pt}

\begin{table}[h!]
    \centering

    \caption{Performance comparison of augmentation methods.}
          \vspace{-6pt}
    \label{tab:results_comparison}
    \renewcommand{\arraystretch}{1.2}
    \resizebox{0.95\linewidth}{!}{%
        \begin{tabular}{l|c|c|c|c}
            \toprule
            \textbf{Method} & \textbf{K.A} \daugshifted & \textbf{K.R} \daugshifted & \textbf{Avg} \daugshifted & \textbf{HARS} \daugshifted \\  
            \midrule

            \rowcolor{backblue!80}
            \textbf{\aug} & \textbf{38.82} & \textbf{35.78} & \textbf{37.30} & \textbf{79.04} \\  

            \rowcolor{gray!10}
            \multicolumn{5}{c}{\fontsize{10}{12}\selectfont \textit{\textbf{Augmentation for Text}}} \\
            Knowledge-Aware   & \underline{20.29} & 34.86 & \underline{27.58} & \underline{69.43} \\
            Knowledge-Agnostic & 15.60 & \underline{35.71} & 25.65 & 63.54 \\

            \rowcolor{gray!10}
            \multicolumn{5}{c}{\fontsize{10}{12}\selectfont \textit{\textbf{Augmentation for Images}}} \\
            Knowledge-Aware   & 18.33 & 34.02 & 26.17 & 66.48 \\
            Knowledge-Agnostic & 18.33 & 32.09 & 25.21 & 64.71 \\

            \bottomrule
        \end{tabular}%
    }
\end{table}

\vspace{-5pt}

\subsection{\label{sec:augmentation}Comparison with general augmentation methods}

This section validates our claim from \autoref{sec:aug} that \aug is superior to general augmentation methods. As shown in Table~\ref{tab:results_comparison}, \aug outperforms general augmentation methods across all metrics, notably achieving improvements of $18.53\uparrow$ in K.A, $9.72\uparrow$ in Avg and $9.61\uparrow$ in HARS over the best baseline.  Since both \aug and general augmentation for text utilize GPT-4o, the substantial performance gap proves that \aug’s effectiveness stems from its superior design rather than mere distillation from GPT-4o.


\section{Conclusion}


In this work, we propose \method, a synergistic method centered on knowledge-oriented controls that addresses the critical trade-off between injecting new knowledge and preserving existing knowledge. Specifically, \method automatically converts each piece of knowledge into a more profound and structured format, ensuring the model accurately learns and adapts to new knowledge. Simultaneously, it minimizes interference with previous knowledge by initializing an adapter with null space that stores the covariance matrix of previous knowledge, enabling powerful retention. \method's robust performance is architecture-agnostic and exhibits enhanced superiority on larger-scale LMMs.



\section*{Impact Statement}

This paper presents work whose goal is to advance the field of 
Machine Learning. There are many potential societal consequences 
of our work, none which we feel must be specifically highlighted here.

\section*{Acknowledgement}

This research is supported by Smart-Grid National Science and Technology Major Project (Grant No. 2025ZD0805500).


\nocite{langley00}

\bibliography{main}
\bibliographystyle{icml2026}

\newpage
\appendix
\onecolumn

\section{The Use of Large Language Models in \method}\label{appendix:Use}

In this section, we elaborate on the precise role of large language models within \method, as detailed below.

\textbf{Usage 1: \traindata's construction.} In \autoref{sec:aug} and \autoref{appendix:ko_augmentation}, we use GPT-4o to generate multi-rounds dialogue data, summary content of original knowledge, and quadruplets $(Q, A, S, H)$ data, which is in line with current scientific research standards

\textbf{Usage 2: Knowledge Retention Evaluation.} In \autoref{sec:Experiment}, we employ MIA-Bench, MMDU, MathVista, and MathVision, whose evaluation requires large language models as judges, a practice consistent with current research standards.

\textbf{Usage 3: Paper grammar polishing.} The initial draft of the paper was written by humans and later refined for grammar using large language models, a common practice in contemporary research.

\section{More details about Setup and Experimental Operation}\label{appendix:setup}

\subsection{Knowledge Adaptation Evaluation}

Our knowledge adaptation evaluation completely follows the settings of EVOKE. Below we will provide a brief introduction: This paper introduces EVOKE \cite{jiang2025when}, a new benchmark to evaluate how well Large Multimodal Models can learn evolving knowledge without forgetting their original capabilities. It reveals the limitations of current methods in knowledge adaptation and the severity of catastrophic forgetting. The study further shows that knowledge augmentation and continual learning are promising solutions, providing a framework for future research.

\subsection{Knowledge Retention Evaluation}

We evaluate fine-tuned LMMs' knowledge retention capabilities on 12 benchmarks across 7 capability dimensions. And we follow the settings of VLMEvalKit \cite{duan2024vlmevalkit} to evaluate these benchmarks, and the following is an introduction:

\textbf{Comprehensive Evaluation:} \ding{182} \textbf{MME} \cite{fu2023mme} provides a holistic evaluation of LMMs' perception and cognition across 14 tasks. Its key feature is the use of carefully crafted instruction-answer pairs, which facilitates a straightforward assessment without the need for specialized prompt engineering. \ding{183} \textbf{MMBench} \cite{liu2023mmbench} is a cross-lingual benchmark for comprehensively evaluating LMMs. It features over 3,000 bilingual multiple-choice questions spanning 20 skill dimensions, from visual recognition to abstract reasoning.

\textbf{Optical
 Character Recognition:} \ding{182} \textbf{SEEDBench2\_Plus} \cite{li2024seed2plus} benchmarks LMMs on interpreting text-rich visuals (\eg charts, web layouts). It uses 2,300 multiple-choice questions to test reasoning capabilities where integrating textual and visual information is essential. \ding{183} \textbf{OCRVQA} \cite{mishra2019ocr} is a benchmark for evaluating a model's ability to answer questions by reading text within images. It focuses on tasks where textual information is essential, requiring tight integration of visual perception and OCR.

\textbf{Multidisciplinary Reasoning:} \ding{182} \textbf{ScienceQA} \cite{lu2022scienceqa} evaluates scientific reasoning through a large-scale multimodal benchmark; it features curriculum-based questions with diagrams and provides lectures and explanations for each question to encourage complex reasoning. \ding{183} \textbf{MMMU} \cite{mmmu} evaluates LMMs on college-level, multimodal questions requiring expert knowledge. The benchmark includes 11,500 questions from six disciplines, utilizing 30 image formats to test complex, subject-specific reasoning.

\textbf{Instruction Following:} \ding{182} \textbf{MIA-Bench} \cite{qian2024mia} is a targeted benchmark that measures how precisely LMMs can follow complex and multi-layered instructions. It consists of 400 distinct image-prompt combinations engineered to test a model's ability to comply with detailed and nuanced directives.

\textbf{Multi-Turn Multi
Image Dialog Understanding:} \ding{182} \textbf{MMDU} \cite{mmdu} evaluates LMMs in multi-image, multi-turn conversational scenarios. It specifically assesses a model's capacity for contextual understanding, temporal reasoning, and maintaining coherence throughout extended interactions.

\textbf{Mathematical Reasoning:} \ding{182} \textbf{MathVista} \cite{mathvista} benchmarks the mathematical reasoning of foundation models in visual contexts. It aggregates 6,141 problems from 31 datasets, requiring detailed visual analysis and compositional logic for solution. \ding{183} \textbf{MathVision} \cite{math_vision} provides a challenging dataset of 3,040 visually-presented problems from math competitions. Categorized into 16 mathematical areas and five difficulty tiers, it offers a structured evaluation of advanced reasoning in LMMs.

\textbf{Hallucination:} \ding{182} \textbf{HallusionBench} \cite{hallusionbench} diagnoses hallucination and illusion in LMMs' visual interpretations. It employs 346 images and 1,129 structured questions to quantitatively analyze the causes of inaccurate or inconsistent model responses. \ding{183} \textbf{POPE} \cite{pope} evaluates object hallucination in LMMs—the tendency to describe non-existent objects. It uses a polling-based questioning strategy to reliably measure this tendency.

\subsection{Evaluation Protocol}

To evaluate performance on open-domain question answering tasks, two key metrics are employed: \textbf{Cover Exact Match (CEM)} and \textbf{F1-Score (F1)}. The \textbf{CEM} metric determines whether the ground truth answer is fully contained within the model's prediction \cite{xu2023lvlm,Jiang2025MINEDPA,jia2026benchmarking,Du2025MMKEBenchAM}. It is defined by the equation:
\[
CEM = \begin{cases} 
1, & y_q \subseteq \hat{Y} \\
0, & \text{otherwise}
\end{cases}
\]
where $y_q$ represents the ground truth answer and $\hat{Y}$ is the text generated by the model.

The \textbf{F1-Score}, on the other hand, assesses the word-level overlap between the predicted and ground truth answers, providing a harmonic mean of \textbf{Precision} and \textbf{Recall} \cite{chan2024rqrag,Peng2025CanVI,fu2026mmku}. Given the ground truth as a set of words \( \mathcal{W}(y_q) = \{ y_1, \dots, y_m \} \) and the model's prediction as \( \mathcal{W}(\hat{Y}) = \{ \hat{y}_1, \dots, \hat{y}_n \} \), the number of common words is calculated as \( \mathcal{U}(\hat{Y}, y_q) = \sum_{t \in \mathcal{W}(y_q)} \mathbf{1}[t \in \mathcal{W}(\hat{Y})] \), where \( \mathbf{1}[\cdot] \) is the indicator function. 

Based on this, Precision is defined as the fraction of predicted words that are correct, $\mathcal{P}(\hat{Y}, Y) = \frac{\mathcal{U}(\hat{Y}, y_q)}{|\mathcal{W}(\hat{Y})|}$, while Recall represents the fraction of ground truth words that were successfully predicted, $\mathcal{R}(\hat{Y}, Y) = \frac{\mathcal{U}(\hat{Y}, y_q)}{|\mathcal{W}(Y)|}$.

\subsection{Baseline Methods}

In this section, we provide a brief introduction to the baseline method, as follows:

\noindent \textbf{EWC:} This seminal continual learning work \cite{kirkpatrick2017EWC} introduces Elastic Weight Consolidation (EWC) to mitigate catastrophic forgetting. EWC slows updates to parameters important for prior tasks by imposing a quadratic constraint based on the Fisher Information Matrix, elastically preserving old knowledge during new learning.

\noindent \textbf{LwF:} This work proposes Knowledge Distillation to mitigate catastrophic forgetting \cite{li2017lwf}. The method preserves knowledge by ensuring the new model's predictions on new data align with the old model's outputs, achieving data-free continual learning through output consistency.

\noindent \textbf{LoRA:} This highly efficient method, LoRA \cite{hu2022lora}, fine-tunes models by training only small, injected low-rank matrices while keeping the original weights frozen. This approach reduces computational costs significantly and helps mitigate catastrophic forgetting.

\noindent \textbf{OLoRA:} This work proposes an orthogonal subspace-based method for continual learning \cite{wang2023orthogonal}. It allocates independent, orthogonal parameter subspaces for each task, constraining updates to prevent interference and mitigate catastrophic forgetting via an elegant geometric solution.

\noindent \textbf{MoELoRA:} The method in \cite{MoELoRA2024} combines MoE with contrastive learning for PEFT. It specializes experts for different data types and uses contrastive objectives to guide expert collaboration, achieving parameter-efficient fine-tuning that reduces catastrophic forgetting.

\noindent \textbf{SEFE:} The method in \cite{chen2025sefe} tackles multimodal catastrophic forgetting by separately addressing two types: superficial forgetting of style and essential forgetting of knowledge. A tailored training strategy preserves essential knowledge during continual instruction learning.

\noindent \textbf{CIA:} The method \cite{qiao2024large} addresses the stability-plasticity trade-off by introducing an automated balancing coefficient derived from Taylor expansion of the loss function. This mechanism dynamically weighs gradients and parameters to minimize knowledge interference, while a semantic-driven parameter allocation strategy decides whether to reuse or expand parameters based on instruction similarity.

The implementation of the above baseline methods refer to its open source code or integrated code repository \cite{chen2024coin,guo2025mcitlib}.

\subsection{Training Parameters about \method}

We have displayed some training parameter settings, as shown in Table~\ref{tab:Hyperparameter}.

\begin{table}[h!]
\caption{Hyperparameter settings for the model training on LLaVA-v1.5 {\small(7B)}, LLaVA-v1.5 {\small(13B)} and Qwen2.5-VL {\small(7B)}.}
\label{tab:Hyperparameter}
\centering
\setlength{\tabcolsep}{4pt} 
\resizebox{0.8\linewidth}{!}{
    \begin{tabular}{ccccc}
    \toprule

    \rowcolor{gray!10}
    \multicolumn{5}{c}{\fontsize{10}{12}\selectfont \textit{\textbf{LLaVA-v1.5 {\small(7B)}}}} \\	

    \textbf{Rank} & \textbf{Optimizer} & \textbf{Deepspeed} & \textbf{Epochs} & \textbf{Vision Select Layer} \\
    235 & AdamW & Zero3 & 6 & -2 \\


    \textbf{Weight Decay} & \textbf{Warmup Ratio} & \textbf{LR Schedule} & \textbf{Learning Rate} & \textbf{Batch Size} \\
    0 & 0.03 & cosine decay & $2 \times 10^{-4}$ & 54 \\
    \midrule

    \rowcolor{gray!10}
    \multicolumn{5}{c}{\fontsize{10}{12}\selectfont \textit{\textbf{LLaVA-v1.5 {\small(13B)}}}} \\	

    \textbf{Rank} & \textbf{Optimizer} & \textbf{Deepspeed} & \textbf{Epochs} & \textbf{Vision Select Layer} \\
    235 & AdamW & Zero3 & 6 & -2 \\
    

    \textbf{Weight Decay} & \textbf{Warmup Ratio} & \textbf{LR Schedule} & \textbf{Learning Rate} & \textbf{Batch Size} \\
    0 & 0.03 & cosine decay & $2 \times 10^{-4}$ & 32 \\
    \midrule

    \rowcolor{gray!10}
    \multicolumn{5}{c}{\fontsize{10}{12}\selectfont \textit{\textbf{Qwen2.5-VL {\small(7B)} }}} \\	

    \textbf{Rank} & \textbf{Optimizer} & \textbf{Deepspeed} & \textbf{Epochs} & \textbf{Image Max Pixels} \\
    274 & AdamW & Zero3 & 6 & 262144 \\


    \textbf{Grad Accum Steps} & \textbf{Warmup Ratio} & \textbf{LR Schedule} & \textbf{Learning Rate} & \textbf{Batch Size} \\
    8 & 0.1 & cosine decay & $2 \times 10^{-4}$ & 24 \\

    \bottomrule
    \end{tabular}
} 
\end{table}

\subsection{Experiment Resources about \method}

All training experiments were conducted using 4 NVIDIA H100 GPUs (each with 96 GiB memory). All evaluation experiments were performed on systems equipped with 4 NVIDIA A100 PCIe GPUs (each with 40 GiB memory).

\section{Proof of \method}\label{appendix:proof}

In Section \autoref{sec:con}, \con initializes the LoRA's low-rank matrix $A$ within the null space of the covariance matrix $C$, which represents prior knowledge and capabilities. This claim is the premise for \con's effectiveness, which we formally prove in Theorem~\ref{the:theorem1}. In \method, the LoRA's low-rank matrix $A$ is frozen, and only matrix $B$ is fine-tuned during the process. We demonstrate in Theorem~\ref{the:theorem2} why this operation minimizes interference with prior knowledge and capabilities during fine-tuning. 

Theorem~\ref{the:theorem2} extends Theorem~\ref{the:theorem1}: as long as Theorem~\ref{the:theorem1} ensures that matrix $A$ lies in the null space of the covariance matrix $C$, the final output of each layer, $W^*X$, remains approximately equal to $W_0 X$, regardless of how the parameters of matrix $B$ are adjusted.

\begin{theorem}\label{the:theorem1}
Let $U_{\text{null}}^T$, $W_0$, $A$ be the approximate null space of the model's covariance matrix composed of input activations in linear layers, the pre-training weights of the model, and the low rank matrix in LoRA, respectively. 
\end{theorem}

\textit{Proof.} We aim to prove that under the assumption that $W_0$ is full-rank, the column space of $A$ forms a subset of the column space of $U_{\text{null}}^T$, which means $\mathrm{Col}(A) = \mathrm{Col}(U^T_{\text{null}})$.

Step 1: Based on the definition in Section \autoref{sec:con}: \\

\vspace{-15pt}

\begin{equation}
A = \sqrt{\Sigma^*} {(V^*)}^T
\end{equation}

\vspace{-5pt}

Since $\Sigma^*$ is a diagonal matrix containing singular values, it only scales the columns of ${(V^*)}^T$ without changing their span. Therefore, the column space of matrix $A$ is identical to the column space of ${(V^*)}^T$:

\vspace{-15pt}

\begin{equation}
\mathrm{Col}(A) = \mathrm{Col}({(V^*)}^T)
\end{equation}

Step 2: Based on the SVD of $W_0 U_{\text{null}} U_{\text{null}}^T$ in Section \autoref{sec:con}: \\

\vspace{-15pt}

\begin{equation}
W_0 U_{\text{null}} U_{\text{null}}^T = U^* \Sigma^* {(V^*)}^T
\end{equation}

$V^*$ represents the right singular vectors of $W_0 U_{\text{null}} U_{\text{null}}^T$ and spans its row space. Since $U_{\text{null}}$ is orthogonal, $U_{\text{null}} U_{\text{null}}^T$ projects any matrix onto the subspace spanned by $U_{\text{null}}$. Therefore, when $W_0$ is full-rank, the column space of $W_0 U_{\text{null}} U_{\text{null}}^T$ is identical to the column space of $U^T_{\text{null}}$:

\begin{equation}
\mathrm{Col}(V^*) = \mathrm{Col}(W_0 U_{\text{null}} U_{\text{null}}^T) = \mathrm{Col}(U^T_{\text{null}})
\end{equation}

Step 3: Combining the content of steps 1 and 2: \\

\vspace{-15pt}

\begin{equation}
\mathrm{Col}(A) = \mathrm{Col}(V^*) =\mathrm{Col}(U^T_{\text{null}})
\end{equation}

Thus, the column space of $A$ is identical to the column space of $U^T_{\text{null}}$, completing the proof.

\begin{theorem}\label{the:theorem2}
Let $X^{(l)}$, $W_0^{(l)}$, and $W^{*^{(l)}}$ denote the input activations from pre-trained knowledge, the initial weight matrix of the $l$-th layer before fine-tuning, and the weight matrix of the $l$-th layer after fine-tuning, respectively, for the $l$-th layer of the LMM.
\end{theorem}

\textit{Proof.} We aim to prove that the output of the $l$-th layer remains approximately unchanged after fine-tuning with \method, \eg

\begin{equation}
    W^{*^{(l)}} X^{(l)} \approx W_0^{(l)} X^{(l)},
\end{equation}

In KORE, the fine-tuned output at the $l$-th layer is defined as:

\begin{equation}
W^{*(l)} = W_0^{(l)} - B^{(l)} A^{(l)} + B^{*(l)} A^{(l)}.
\end{equation}

Based on $A^{(l)} X^{(l)} \approx \mathbf{0}$ from Section \autoref{sec:con}, we have: \\

\vspace{-15pt}

\begin{equation}
W^{*(l)} X^{(l)} \approx W_0^{(l)} X^{(l)}.
\end{equation}

The output thus remains approximately unchanged, ensuring that the fine-tuning minimally alters the pre-trained knowledge. This concludes the proof.

\section{More details about analysis of ability to capture knowledge}\label{appendix:evoke_presentation}

\subsection{Detailed experimental results for capture knowledge}

\begin{table*}[h!]
    \caption{The detailed numbers and more results of the experiment in Figure~\ref{fig:capture_know}}
    \label{tab:capture_know}
    \vspace{-6pt}
    \centering
    \resizebox{0.80\linewidth}{!}{%
    \begin{tabular}{c|l|ccccc}
    \toprule
    \multirow{2.5}{*}{\textbf{Test Data}} & \multirow{2.5}{*}{\textbf{Method}} & \multicolumn{5}{c}{\textbf{Discarded Ranks}} \\
    \cmidrule{3-7}
    & & \textbf{128} & \textbf{256} & \textbf{512} & \textbf{1024} & \textbf{1536} \\
    \midrule
    \multirow{6}{*}{MME} & Plain SVD & 1492.95 & 1487.28 & 1318.18 & 1169.87 & 744.03  \\
    & ASVD (with 256 MME samples) & 1490.14 & 1476.02 & 1488.48 & 1425.41 & 1239.74  \\
    & CO-SVD (with 256 MME samples) & 1498.17 & 1511.25 & 1514.43 & 1486.81 & \textbf{1458.36}  \\
    & CO-SVD (with 32 MME samples) & 1508.90 & 1512.90 & 1507.78 & 1498.81 & 1341.82  \\
    & CO-SVD (with 512 MME samples) & 1507.42 & 1516.68 & 1505.33 & 1460.32 & 1449.82   \\
    & CO-SVD (with 256 ScienceQA samples) & 1486.51 & 1492.65 & 1478.73 & 1419.61 & 1300.89  \\
    \midrule
    \multirow{6}{*}{ScienceQA} & Plain SVD & 67.13 & 66.85 & 65.59 & 50.41 & 0.73  \\
    & ASVD (with 256 ScienceQA samples) & 67.63 & 66.95 & 66.75 & 62.38 & 49.14  \\
    & CO-SVD (with 256 ScienceQA samples) & 67.19 & 67.16 & 67.62 & 67.61 & \textbf{66.76}  \\
    & CO-SVD (with 32 ScienceQA samples) & 67.48 & 66.77 & 66.97 & 66.61 & 64.58  \\
    & CO-SVD (with 512 ScienceQA samples) & 67.08 & 67.00 & 67.40 & 66.91 & 66.27   \\
    & CO-SVD (with 256 MME samples) & 67.74 & 67.49 & 67.53 & 65.69 & 62.43  \\
    \bottomrule
    \end{tabular}%
    }
\end{table*}

Table~\ref{tab:capture_know} presents detailed data and additional results from the experiment illustrated in Figure~\ref{fig:capture_know}. The results indicate that the number of sampled data points has only a limited influence. When the smallest 1536 ranks are discarded, performance with 512 samples is slightly lower than with 256 samples; using 32 samples leads to a more noticeable decline compared to 256 samples, yet still significantly outperforms both Plain SVD and ASVD \cite{yuan2023asvd}. This suggests that even a small number of samples is sufficient to capture essential knowledge into the covariance matrix.

Furthermore, using test-specific samples allows for better performance after discarding a large number of ranks. For instance, when discarding 1536 ranks, CO-SVD (with 256 MME samples) outperforms CO-SVD (with 256 ScienceQA samples) on the MME, while CO-SVD (with 256 ScienceQA samples) surpasses CO-SVD (with 256 MME samples) on ScienceQA. This demonstrates that CO-SVD effectively captures dataset-specific knowledge and preserves structural features in the covariance matrix, enabling knowledge-oriented constraints and resulting in powerful retention.

\subsection{Covariance Visualization Results}

\vspace{-5pt}

\begin{figure}[th]
\centering
\includegraphics[width=0.6\textwidth]{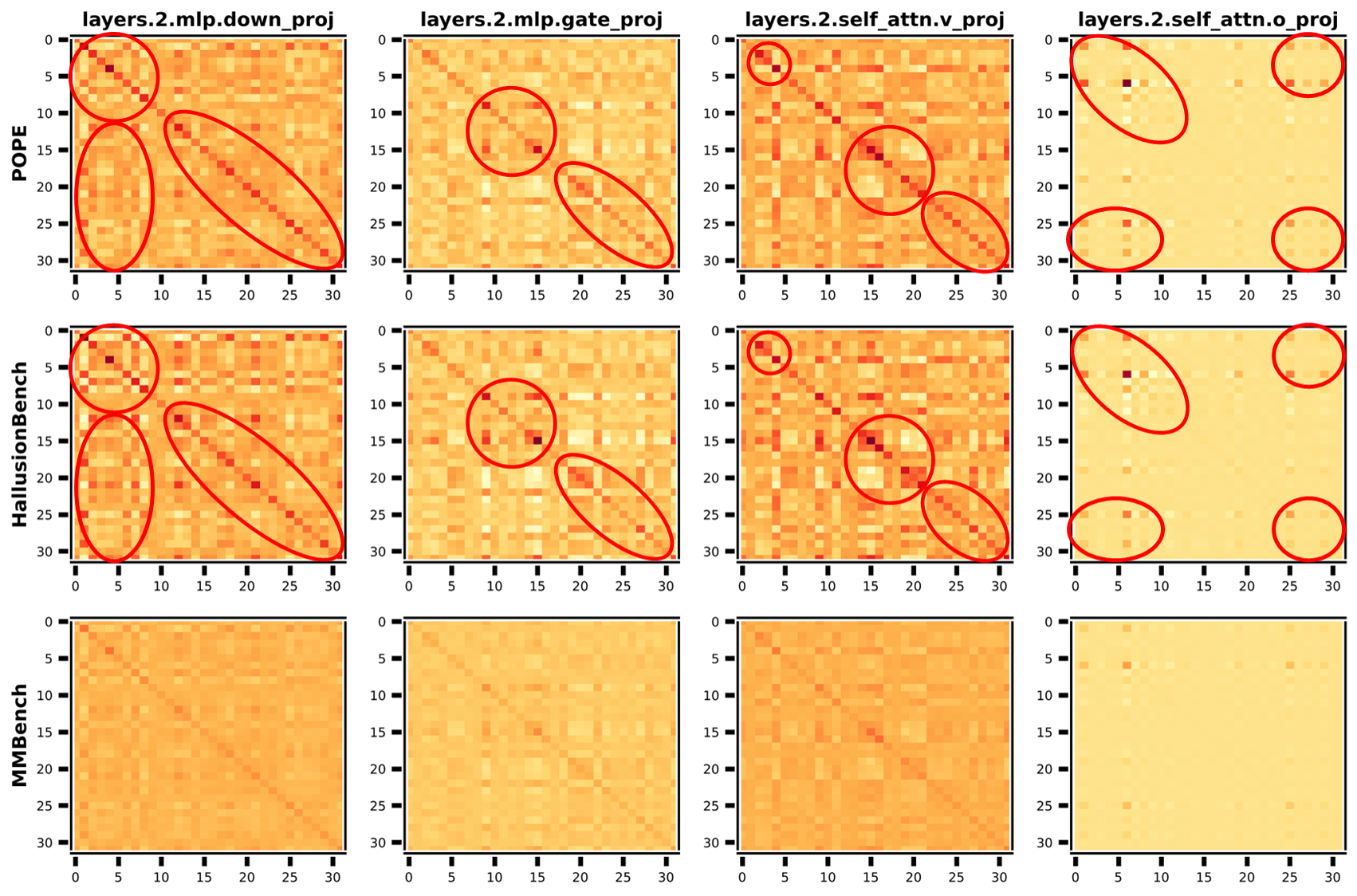}
\vspace{-8pt}
  \caption{Covariance matrix visualization for ``mlp.down\_proj'', ``mlp.gate\_proj'',``self\_attn.v\_proj''and ``self\_attn.o\_proj'' 
  weights in the \textbf{2-th layer} on \textbf{POPE, HallusionBench and MMBench}.}
  \label{fig:HallusionBench2}
\end{figure}

\vspace{-5pt}

\begin{figure}[th]
\centering
\includegraphics[width=0.6\textwidth]{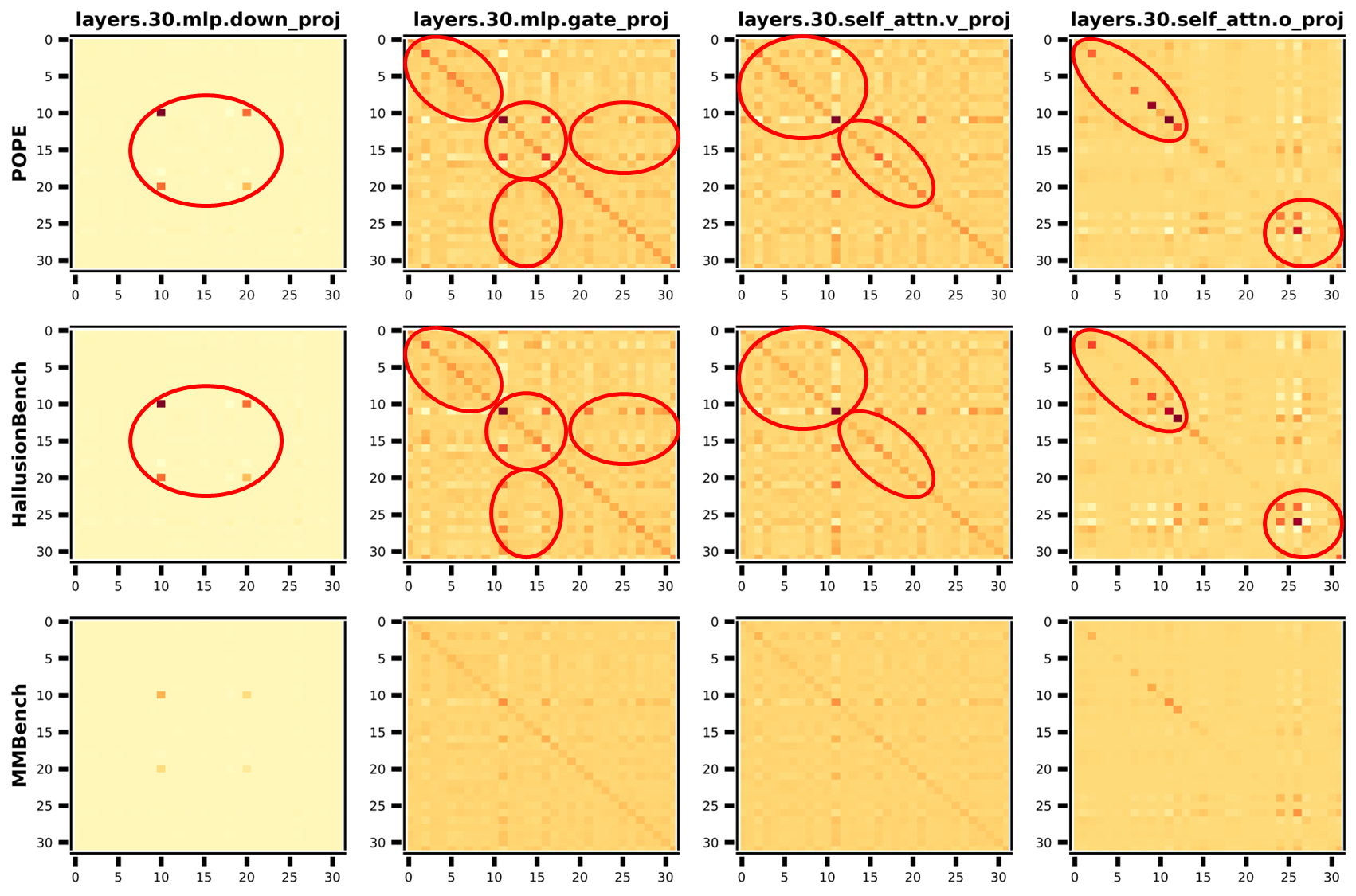}
\vspace{-8pt}
  \caption{Covariance matrix visualization for ``mlp.down\_proj'', ``mlp.gate\_proj'',``self\_attn.v\_proj''and ``self\_attn.o\_proj'' 
  weights in the \textbf{30-th layer} on \textbf{POPE, HallusionBench and MMBench}.}
  \label{fig:HallusionBench30}
\end{figure}

In Figures~\ref{fig:HallusionBench2} and~\ref{fig:HallusionBench30}, we further provide visualizations of the covariance matrices collected from the POPE, HallusionBench, and MMBench tasks. Due to the high and uninformative original dimensionality of 4096 or 11088, we downsampled the covariance matrices to 32×32 and visualized their heatmaps. We present activations prior to various linear weights, including ``mlp.down\_proj'', ``mlp.gate\_proj'',``self\_attn.v\_proj''and ``self\_attn.o\_proj'' from both \textbf{layer 2} and \textbf{layer 30}. The results show that heatmaps from \textbf{POPE} and \textbf{HallusionBench} (both hallucination tasks) share certain similar patterns (highlighted with red circles) not observed in heatmaps from \textbf{MMBench}. This indicates that the activated covariance matrices exhibit distinct patterns when inputs from different tasks are processed by the LMMs. These visualizations empirically support that covariance matrix patterns can characterize the triggered task. We leverage such patterns to guide the decomposition of pre-trained weights in LMMs, obtaining initialized adapters enriched with more informative knowledge.

\section{More experimental results about \method}\label{appendix:results}

\subsection{More Main Results}

Regarding the experiment in Figure~\ref{fig:know_types} in \autoref{sec:detailed}, we have supplemented Table~\ref{table:main_results_knowledge_type} with detailed numerical performance of all methods on fine-grained knowledge types for readers' reference.

\vspace{-10pt}

\begin{table}[th]
\captionsetup{position=top}
\setlength{\abovecaptionskip}{3pt}
\setlength{\belowcaptionskip}{6pt}
\setlength{\textfloatsep}{6pt}
\caption{\textbf{Performance comparison between \method and baseline methods on fine-grained knowledge types with LLaVA-v1.5 {\small(7B)}.} PO: Politics; SP: Sports; BU: Business; HE: Health; CE: Celebrity; FI: Film; AL: Album; WR: Written Work.}
\centering
\renewcommand{\arraystretch}{1.3}
\resizebox{1\linewidth}{!}{
\begin{tabular}{l|cc|cc|cc|cc|cc|cc|cc|cc|cc|cc}
\toprule
\multirow{4}{*}{\textbf{Method}}
& \multicolumn{10}{c|}{\textbf{News}}
& \multicolumn{10}{c}{\textbf{Entity}} \\
\cmidrule{2-21}
& \multicolumn{2}{c|}{\textbf{Avg}}
& \multicolumn{2}{c|}{\textbf{PO}}
& \multicolumn{2}{c|}{\textbf{SP}}
& \multicolumn{2}{c|}{\textbf{BU}}
& \multicolumn{2}{c|}{\textbf{HE}}
& \multicolumn{2}{c|}{\textbf{Avg}}
& \multicolumn{2}{c|}{\textbf{CE}}
& \multicolumn{2}{c|}{\textbf{FI}}
& \multicolumn{2}{c|}{\textbf{AL}}
& \multicolumn{2}{c}{\textbf{WR}} \\
\cmidrule{2-21}
& \textbf{CEM} \daugshifted & \textbf{F1} \daugshifted
& \textbf{CEM} \daugshifted & \textbf{F1} \daugshifted
& \textbf{CEM} \daugshifted & \textbf{F1} \daugshifted
& \textbf{CEM} \daugshifted & \textbf{F1} \daugshifted
& \textbf{CEM} \daugshifted & \textbf{F1} \daugshifted
& \textbf{CEM} \daugshifted & \textbf{F1} \daugshifted
& \textbf{CEM} \daugshifted & \textbf{F1} \daugshifted
& \textbf{CEM} \daugshifted & \textbf{F1} \daugshifted
& \textbf{CEM} \daugshifted & \textbf{F1} \daugshifted
& \textbf{CEM} \daugshifted & \textbf{F1} \daugshifted \\
\midrule


Full-FT & \underline{21.35} & 16.34 & 12.92 & 10.99 & 22.49 & 20.88 & \underline{27.31} & 20.95 & 19.84 & 16.47 & \underline{14.37} & 13.88 & 13.11 & 16.93 & \underline{12.39} & 13.16 & \underline{12.17} & 7.66 & \underline{20.34} & 8.43 \\
LoRA & 17.72 & 19.42 & 10.54 & 12.96 & 19.11 & 21.50 & 20.66 & 24.03 & 17.81 & 23.76 & 12.51 & 17.09 & 12.20 & 21.19 & 10.57 & 15.82 & 10.72 & 8.72 & 18.64 & 12.94 \\

\midrule
Replay & 13.98 & 19.43 & 7.61 & 13.16 & 15.96 & 20.69 & 16.05 & 22.40 & 15.38 & 24.21 & 8.48 & 16.39 & 9.40 & 18.78 & 10.34 & 15.60 & 3.77 & 10.79 & 4.55 & 8.23 \\
EWC & 17.86 & 21.10 & 10.45 & 14.81 & 19.83 & 23.02 & 19.00 & 24.57 & 17.41 & 23.88 & 12.88 & 17.58 & \underline{14.53} & 22.07 & 12.16 & 16.91 & 10.72 & 8.13 & 15.25 & 17.69 \\
LwF & 17.05 & 21.43 & 9.62 & 13.99 & 19.83 & 23.66 & 18.63 & 25.82 & 19.03 & 26.20 & 11.88 & 18.40 & 12.45 & 21.64 & \underline{12.39} & 17.01 & 9.28 & 11.11 & 10.17 & 17.10 \\
MoELoRA & 9.23 & 14.86 & 3.39 & 8.72 & 6.77 & 11.77 & 12.36 & 18.92 & 10.53 & 20.60 & 3.40 & 9.28 & 2.95 & 10.32 & 4.43 & 8.96 & 3.19 & 5.22 & 10.17 & 14.07 \\
O-LoRA & 9.21 & 14.68 & 3.67 & 8.52 & 7.01 & 12.23 & 12.55 & 18.98 & 11.74 & 20.68 & 3.40 & 9.22 & 3.10 & 10.51 & 4.20 & 8.28 & 3.19 & 5.35 & 8.47 & 12.37 \\

SEFE & 21.12 & \underline{26.93} & \underline{14.85} & \underline{20.94} & \underline{22.61} & \underline{26.60} & 23.43 & \underline{31.88} & \underline{20.65} & \underline{33.27} & 10.84 & \underline{21.05} & 13.87 & \underline{25.58} & 8.30 & 14.40 & 6.67 & \underline{17.41} & 15.25 & \underline{26.27} \\

CIA & 17.37 & 21.71 & 10.27 & 14.85 & 20.56 & 23.53 & 18.08 & 25.46 & 17.41 & 25.39 & 11.35 & 18.70 & 12.45 & 23.21 & 11.59 & \underline{17.15} & 9.57 & 8.72 & 11.86 & 11.72 \\

\midrule
\rowcolor{backblue!80}
\textbf{\method} & \textbf{34.74} & \textbf{42.96} & \textbf{23.83} & \textbf{32.31} & \textbf{46.19} & \textbf{50.38} & \textbf{34.69} & \textbf{45.74} & \textbf{33.20} & \textbf{45.23} & \textbf{26.17} & \textbf{39.39} & \textbf{27.79} & \textbf{42.61} & \textbf{26.93} & \textbf{34.05} & \textbf{16.52} & \textbf{29.54} & \textbf{28.81} & \textbf{43.05} \\
\bottomrule
\end{tabular}
}
\label{table:main_results_knowledge_type}
\end{table}

\subsection{More Results on LMM Scales and Architectures}\label{appendix:More_Results_Scales}

Regarding the experiment in \autoref{sec:scales}, we supplement the detailed results of knowledge injection in Tables~\ref{tab:lmm_scales_architectures_overall}, ~\ref{table:scale_per_benchmark} and ~\ref{table:scale_knowledge_type}.

\begin{table*}[h]
\caption{Performance comparison between \method and baseline methods with LLaVA-v1.5 {\small(13B)} and Qwen2.5-VL {\small(7B)}.}
  \label{tab:lmm_scales_architectures_overall}
  \vspace{-6pt}
  \centering
  \renewcommand{\arraystretch}{1.2} 
  \resizebox{\textwidth}{!}{%
    \begin{tabular}{l|cc|ccccccc|cccc}
      \toprule
      \multirow{2.5}{*}{\textbf{Method}}
      & \multicolumn{2}{c|}{\textbf{\dataset}}
      & \multirow{2.5}{*}{\textbf{COM} \daugshifted}
      & \multirow{2.5}{*}{\textbf{OCR} \daugshifted}
      & \multirow{2.5}{*}{\textbf{M-DIS} \daugshifted}
      & \multirow{2.5}{*}{\textbf{INS} \daugshifted}
      & \multirow{2.5}{*}{\textbf{M-IDU} \daugshifted}
      & \multirow{2.5}{*}{\textbf{MAT} \daugshifted}
      & \multirow{2.5}{*}{\textbf{HAL} \daugshifted}
      & \multirow{2.5}{*}{\textbf{K.A} \daugshifted}
      & \multirow{2.5}{*}{\textbf{K.R} \daugshifted}
      & \multirow{2.5}{*}{\textbf{Avg} \daugshifted}
      & \multirow{2.5}{*}{\textbf{HARS} \daugshifted} \\
      \cmidrule{2-3} 
      & \textbf{CEM} \daugshifted & \textbf{F1}\daugshifted  & & & & & & & & & & & \\

\midrule

      \multicolumn{14}{c}{\fontsize{10}{12}\selectfont \textit{\textbf{LLaVA-v1.5 {\small(13B)}}}} \\

      \midrule
      \rowcolor{gray!10}
      Vanilla &5.33&10.07&66.86&51.12&52.70&66.04&33.93&19.64&56.77&7.70&49.51&28.60& --- \\
       \midrule
      LoRA &\underline{16.26}&\underline{22.83}&45.28&32.58&43.72&23.26&17.43&15.82&38.08&\underline{19.55}&32.64&26.09&63.15  \\
      Replay &12.05&20.21&\textbf{61.65}&\textbf{47.51}&\underline{48.42}&\underline{61.04}&\underline{24.62}&\underline{19.55}&\textbf{54.16}&16.13&\textbf{45.69}&\underline{30.91}&\underline{66.73}  \\
\rowcolor{backblue!80}
      
     \textbf{\method} &\textbf{32.89}&\textbf{44.47}&\underline{59.35}&\underline{45.96}&\textbf{51.39}&\textbf{65.10}&\textbf{26.84}&\textbf{20.31}&\underline{40.52}&\textbf{38.68}&\underline{45.35}&\textbf{42.01}&\textbf{85.46} \\

      \midrule

\multicolumn{14}{c}{\fontsize{10}{12}\selectfont \textit{\textbf{Qwen2.5-VL {\small(7B)}}}} \\

      \midrule
      \rowcolor{gray!10}
      Vanilla &9.34&15.33&81.18&70.32&65.35&78.46&61.25&47.69&66.96&12.34&66.89&39.61& --- \\
       \midrule
      LoRA &\underline{14.56}&14.01&52.54&64.54&22.35&21.39&23.25&13.52&41.38&14.29&38.16&26.22&22.03 \\
      Replay &11.73&\underline{18.51}&\textbf{78.54}&\textbf{69.17}&\underline{65.26}&\underline{70.20}&\textbf{50.72}&\underline{42.74}&\textbf{67.48}&\underline{15.12}&\textbf{63.94}&\underline{39.53}&\underline{30.89} \\
\rowcolor{backblue!80}
      
     \textbf{\method} &\textbf{22.91}&\textbf{31.36}&\underline{56.60}&\underline{67.74}&\textbf{65.48}&\textbf{70.51}&\underline{45.02}&\textbf{43.72}&\underline{58.57}&\textbf{27.14}&\underline{58.31}&\textbf{42.72}&\textbf{67.10} \\
      
      \bottomrule
    \end{tabular}%
  }
\end{table*}

\begin{table*}[ht]
\caption{Performance comparison between \method and baseline methods on fine-grained knowledge retention evaluations with LLaVA-v1.5 {\small(13B)} and Qwen2.5-VL {\small(7B)}.}

\vspace{-6pt}
\centering
\resizebox{1\linewidth}{!}{
\begin{tabular}{l|c c|c c|c c|c|c|c c|c c|c}
\toprule
\multirow{2.5}{*}{\textbf{Method}} & \multicolumn{2}{c|}{\textbf{COM}} & \multicolumn{2}{c|}{\textbf{OCR}} & \multicolumn{2}{c|}{\textbf{M-DIS}} & \multicolumn{1}{c|}{\textbf{INS}} & \multicolumn{1}{c|}{\textbf{M-IDU}} & \multicolumn{2}{c|}{\textbf{MAT}} & \multicolumn{2}{c|}{\textbf{HAL}} & \multirow{2.5}{*}{\textbf{Avg}} \\
\cmidrule{2-13}
& \textbf{MME} \daugshifted & \textbf{MM}\raisebox{1ex}{\tiny\textbf{B}} \daugshifted & \textbf{SEED}\raisebox{1ex}{\tiny\textbf{B2P}} \daugshifted & \textbf{OCR}\raisebox{1ex}{\tiny\textbf{VQA}} \daugshifted & \textbf{SQA} \daugshifted & \textbf{MMMU} \daugshifted &\textbf{MIA}\raisebox{1ex}{\tiny\textbf{B}} \daugshifted & \textbf{MMDU} \daugshifted & \textbf{Math}\raisebox{1ex}{\tiny\textbf{T}} \daugshifted & \textbf{Math}\raisebox{1ex}{\tiny\textbf{I}} \daugshifted & \textbf{POPE} \daugshifted & \textbf{Hall}\raisebox{1ex}{\tiny\textbf{B}} \daugshifted \\
\midrule

      \multicolumn{14}{c}{\fontsize{10}{12}\selectfont \textit{\textbf{LLaVA-v1.5 {\small(13B)}}}} \\	
  
       \rowcolor{gray!10}  Vanilla  &65.33&68.38&42.25&59.99&73.90&31.50&66.04&33.93&27.40&11.88&87.07&26.46&49.51 \\
         \midrule
      LoRA &30.00&60.57&36.93&28.22&69.13&18.30&23.26&17.43&23.90&7.73&71.64&4.52& 32.64      \\ 
      
      Replay
      &\textbf{57.49}&\textbf{65.81}&\underline{40.27}&\textbf{54.75}&\underline{70.94}&\underline{25.90}&\underline{61.04}&\underline{24.62}&\underline{27.00}&\underline{12.11}&\textbf{87.09}&\textbf{21.23}& \textbf{45.69}   \\

      \midrule


\rowcolor{backblue!80}
      \textbf{\method}  &\underline{55.99}&\underline{62.71}&\textbf{40.32}&\underline{51.60}&\textbf{71.97}&\textbf{30.80}&\textbf{65.10}&\textbf{26.84}&\textbf{27.30}&\textbf{13.32}&\underline{79.29}&\underline{18.91}&\underline{45.35}       \\

      \midrule

      \multicolumn{14}{c}{\fontsize{10}{12}\selectfont \textit{\textbf{Qwen2.5-VL {\small(7B)}}}} \\
\rowcolor{gray!10}
              Vanilla  &82.54&79.81&69.61&71.03&72.10&58.60&78.46&61.25&69.70&25.69&86.51&47.42&66.89\\
         \midrule
  
     
      LoRA     &\underline{67.88}&37.20&59.29&69.79&42.30&2.40&21.39&23.25&39.40&13.52&73.73&9.02&38.16   \\
      
      Replay
      &\textbf{75.38}&\textbf{81.70}&\textbf{69.16}&\textbf{69.17}&\underline{85.12}&\textbf{45.40}&\underline{70.20}&\textbf{50.72}&\textbf{63.90}&21.58&\textbf{87.49}&\textbf{47.48}& \textbf{63.94}  \\

      \midrule
      \rowcolor{backblue!80}
      \textbf{\method}      &36.23&\underline{76.98}&\underline{66.80}&\underline{68.69}&\textbf{85.55}&\textbf{45.40}&\textbf{70.51}&\underline{45.02}&\underline{63.10}&\textbf{24.34}&\underline{75.24}&\underline{41.89}&\underline{58.31}   \\

\bottomrule
\end{tabular}

}
\label{table:scale_per_benchmark}
\end{table*}

\textbf{Obs 1 in \autoref{appendix:More_Results_Scales}: \method still achieves superior knowledge retention performance on larger-scale LMM and different model architectures.} As shown in Table~\ref{table:scale_per_benchmark}, on LLaVA-v1.5 {\small(13B)}, \method outperforms Replay on seven benchmarks and achieves comparable overall performance. This result demonstrates \method's potential for superior performance on larger-scale LMM. On Qwen2.5-VL {\small(7B)}, \method surpasses LoRA by $20.15$ in overall performance, demonstrating its ability to maintain superior knowledge retention across different model architectures and confirming its universality and robustness.

\begin{table}[ht]
\captionsetup{position=top}
\setlength{\abovecaptionskip}{3pt}
\setlength{\belowcaptionskip}{6pt}
\setlength{\textfloatsep}{6pt}
\caption{{Performance comparison between \method and baseline methods on fine-grained knowledge types with LLaVA-v1.5 {\small(13B)} and Qwen2.5-VL {\small(7B)}.}}
\centering

\renewcommand{\arraystretch}{1.3}
\resizebox{1\linewidth}{!}{
\begin{tabular}{l|cc|cc|cc|cc|cc|cc|cc|cc|cc|cc}
\toprule
\multirow{4}{*}{\textbf{Method}}
& \multicolumn{10}{c|}{\textbf{News}}
& \multicolumn{10}{c}{\textbf{Entity}} \\
\cmidrule{2-21}
& \multicolumn{2}{c|}{\textbf{Avg}}
& \multicolumn{2}{c|}{\textbf{PO}}
& \multicolumn{2}{c|}{\textbf{SP}}
& \multicolumn{2}{c|}{\textbf{BU}}
& \multicolumn{2}{c|}{\textbf{HE}}
& \multicolumn{2}{c|}{\textbf{Avg}}
& \multicolumn{2}{c|}{\textbf{CE}}
& \multicolumn{2}{c|}{\textbf{FI}}
& \multicolumn{2}{c|}{\textbf{AL}}
& \multicolumn{2}{c}{\textbf{WR}} \\
\cmidrule{2-21}
& \textbf{CEM} \daugshifted & \textbf{F1} \daugshifted
& \textbf{CEM} \daugshifted & \textbf{F1} \daugshifted
& \textbf{CEM} \daugshifted & \textbf{F1} \daugshifted
& \textbf{CEM} \daugshifted & \textbf{F1} \daugshifted
& \textbf{CEM} \daugshifted & \textbf{F1} \daugshifted
& \textbf{CEM} \daugshifted & \textbf{F1} \daugshifted
& \textbf{CEM} \daugshifted & \textbf{F1} \daugshifted
& \textbf{CEM} \daugshifted & \textbf{F1} \daugshifted
& \textbf{CEM} \daugshifted & \textbf{F1} \daugshifted
& \textbf{CEM} \daugshifted & \textbf{F1} \daugshifted \\
\midrule

\rowcolor{gray!10}
      \multicolumn{21}{c}{\fontsize{10}{12}\selectfont \textit{\textbf{LLaVA-v1.5 {\small(13B)}}}} \\

LoRA &\underline{20.15}&\underline{25.10}&\underline{12.65}&\underline{16.17}&\underline{24.79}&\underline{28.69}&\underline{21.77}&\underline{29.51}&\underline{22.27}&\underline{29.09}&\underline{11.99}&\underline{20.34}&\underline{13.72}&\underline{25.26}&\underline{13.18}&\underline{18.04}&\underline{6.67}&{12.18}&\underline{10.17}&{15.87} \\
        Replay &15.04&21.83&8.16&14.41&15.60&21.76&15.87&24.74&18.62&28.74&8.77&18.42&9.45&21.50&10.91&17.16&5.51&\underline{13.38}&\underline{10.17}&\underline{20.97} \\

\midrule
\rowcolor{backblue!80}
\textbf{\method} &\textbf{36.77}&\textbf{46.11}&\textbf{25.39}&\textbf{34.41}&\textbf{47.16}&\textbf{53.39}&\textbf{37.45}&\textbf{50.95}&\textbf{35.22}&\textbf{48.51}&\textbf{28.64}&\textbf{42.67}&\textbf{28.66}&\textbf{44.95}&\textbf{31.02}&\textbf{38.21}&\textbf{22.61}&\textbf{35.43}&\textbf{20.34}&\textbf{33.06} \\

\rowcolor{gray!10}
      \multicolumn{21}{c}{\fontsize{10}{12}\selectfont \textit{\textbf{Qwen2.5-VL {\small(7B)}}}} \\

LoRA  &\underline{17.76}&{14.09}&\underline{12.01}&{7.18}&\underline{17.41}&{17.65}&\underline{22.32}&{17.90}&\underline{19.03}&{17.21}&\underline{11.06}&{13.93}&\underline{8.03}&{15.91}&\underline{21.48}&{14.91}&{8.70}&{10.87}&\underline{16.95}&{11.32} \\

        Replay  &13.45&\underline{18.40}&7.33&\underline{11.09}&14.03&\underline{17.94}&14.58&\underline{22.72}&15.38&\underline{23.72}&9.84&\underline{18.63}&7.16&\underline{17.69}&20.45&\underline{28.00}&\underline{9.28}&\underline{12.97}&\underline{16.95}&\underline{24.89} \\

\midrule
\rowcolor{backblue!80}

\textbf{\method}  
&\textbf{26.93}&\textbf{32.51}&\textbf{17.42}&\textbf{22.75}&\textbf{31.20}&\textbf{35.11}&\textbf{31.00}&\textbf{39.43}&\textbf{33.20}&\textbf{40.49}&\textbf{18.51}&\textbf{30.11}&\textbf{16.11}&\textbf{28.63}&\textbf{26.14}&\textbf{33.20}&\textbf{13.33}&\textbf{25.91}&\textbf{25.42}&\textbf{41.24} \\

\bottomrule
\end{tabular}
}
\label{table:scale_knowledge_type}
\end{table}

\textbf{Obs 2 in \autoref{appendix:More_Results_Scales}: \method achieves comprehensive performance advantages across diverse knowledge types on both larger-scale LMM and different model architectures.} In Table~\ref{table:scale_knowledge_type}, \method achieves the best knowledge adaptation performance across all news and entity types on both LLaVA-v1.5 {\small(13B)} and Qwen2.5-VL {\small(7B)}, significantly outperforming LoRA and Replay. This demonstrates that KORE's effectiveness in new knowledge injection is not constrained by model scale or architecture, highlighting its powerful universality.

\subsection{More Results on Specific Knowledge-Oriented Constraints}\label{appendix:specific}

For the experiment on specific knowledge-oriented constraints in \autoref{sec:detailed}, we have provided detailed results and presented them below.

\begin{table*}[h]
\caption{Performance comparison of specific knowledge-oriented constraints with LLaVA-v1.5 {\small(7B)}.}
    \label{tab:specific_main_appendix}
  \vspace{-6pt}
  \centering
  \renewcommand{\arraystretch}{1.2} 
  \resizebox{\textwidth}{!}{%
    \begin{tabular}{l|cc|ccccccc|cccc}
      \toprule
      \multirow{2.5}{*}{\textbf{Method}}
      & \multicolumn{2}{c|}{\textbf{\dataset}}
      & \multirow{2.5}{*}{\textbf{COM} \daugshifted}
      & \multirow{2.5}{*}{\textbf{OCR} \daugshifted}
      & \multirow{2.5}{*}{\textbf{M-DIS} \daugshifted}
      & \multirow{2.5}{*}{\textbf{INS} \daugshifted}
      & \multirow{2.5}{*}{\textbf{M-IDU} \daugshifted}
      & \multirow{2.5}{*}{\textbf{MAT} \daugshifted}
      & \multirow{2.5}{*}{\textbf{HAL} \daugshifted}
      & \multirow{2.5}{*}{\textbf{K.A} \daugshifted}
      & \multirow{2.5}{*}{\textbf{K.R} \daugshifted}
      & \multirow{2.5}{*}{\textbf{Avg} \daugshifted}
      & \multirow{2.5}{*}{\textbf{HARS} \daugshifted} \\
      \cmidrule{2-3} 
      & \textbf{CEM} \daugshifted & \textbf{F1}\daugshifted  & & & & & & & & & & & \\

      \midrule
\rowcolor{backblue!80}
       \textbf{\method} &\textbf{30.65}&\textbf{41.26}&52.41&\underline{40.98}&48.68&38.54&16.58&\underline{18.59}&51.75&\textbf{35.96}&40.00&37.98&82.81 \\
\midrule
\textbf{\mme} &29.48&39.44&\textbf{56.90}&39.86&47.41&57.06&\textbf{27.70}&17.92&\underline{52.20}&34.46&\textbf{43.03}&38.75&\textbf{85.24} \\
\textbf{\ocr} &29.95&39.75&\underline{52.60}&\textbf{41.47}&\textbf{48.86}&60.10&27.09&18.28&50.15&34.85&42.24&38.54&84.64 \\
\textbf{\matht} &\underline{30.06}&\underline{40.33}&52.40&40.32&\underline{48.57}&\underline{60.30}&\underline{27.69}&\textbf{19.24}&51.57&\underline{35.20}&\underline{42.68}&\textbf{38.94}&\underline{85.16} \\
\textbf{\hall} &29.93&39.98&54.37&36.68&46.50&\textbf{60.71}&26.30&17.42&\textbf{52.67}&34.96&41.86&38.41&84.31 \\

      \bottomrule
    \end{tabular}%
  }
\end{table*}

\textbf{Obs 1 in \autoref{appendix:specific}: \method with specific knowledge-oriented constraints achieves superior comprehensive performance.} In Table~\ref{tab:specific_main_appendix}, \method with specific knowledge-oriented constraints (\eg {MME}, {OCR}\raisebox{1ex}{\tiny{VQA}}, {Math}\raisebox{1ex}{\tiny{T}}, {Hall}\raisebox{1ex}{\tiny{B}}) causes a slight decrease in knowledge adaptation efficacy, it yields a significant increase in knowledge retention performance on INS and M-IDU, resulting in a superior overall performance.

\textbf{Obs 2 in \autoref{appendix:specific}: Specific knowledge-oriented constraints enhance the retention of corresponding knowledge.} In Table~\ref{table:specific_each_benchmark}, specific knowledge-oriented constraints enhance the retention of corresponding knowledge without compromising the retention of other knowledge types. This capability underscores \method 's potential for applications requiring customized knowledge preservation.

\begin{table*}[h!]
\caption{Performance of specific knowledge-oriented constrains on fine-grained knowledge retention evaluations with LLaVA-v1.5 {\small(7B)}.}

\vspace{-6pt}

\centering
\resizebox{1\linewidth}{!}{
\begin{tabular}{l|c c|c c|c c|c|c|c c|c c|c}
\toprule
\multirow{2.5}{*}{\textbf{Method}} & \multicolumn{2}{c|}{\textbf{COM}} & \multicolumn{2}{c|}{\textbf{OCR}} & \multicolumn{2}{c|}{\textbf{M-DIS}} & \multicolumn{1}{c|}{\textbf{INS}} & \multicolumn{1}{c|}{\textbf{M-IDU}} & \multicolumn{2}{c|}{\textbf{MAT}} & \multicolumn{2}{c|}{\textbf{HAL}} & \multirow{2.5}{*}{\textbf{Avg}} \\
\cmidrule{2-13}
& \textbf{MME} \daugshifted & \textbf{MM}\raisebox{1ex}{\tiny\textbf{B}} \daugshifted & \textbf{SEED}\raisebox{1ex}{\tiny\textbf{B2P}} \daugshifted & \textbf{OCR}\raisebox{1ex}{\tiny\textbf{VQA}} \daugshifted & \textbf{SQA} \daugshifted & \textbf{MMMU}\raisebox{1ex}{\tiny\textbf{T}} \daugshifted &\textbf{MIA}\raisebox{1ex}{\tiny\textbf{B}} \daugshifted & \textbf{MMDU} \daugshifted & \textbf{Math}\raisebox{1ex}{\tiny\textbf{T}} \daugshifted & \textbf{Math}\raisebox{1ex}{\tiny\textbf{I}} \daugshifted & \textbf{POPE} \daugshifted & \textbf{Hall}\raisebox{1ex}{\tiny\textbf{B}} \daugshifted \\
\midrule

\rowcolor{backblue!80}
\textbf{\method}  & 49.84 & 54.98 & \textbf{37.73} & \underline{44.24} & \underline{68.06} & 29.30 & 38.54 & 16.58 & \underline{25.10} & \underline{12.09} & 80.99 & 22.51 & 40.00 \\

\midrule
        \textbf{\mme} & \textbf{57.01} & \textbf{56.79} & \underline{37.51} & 42.22 & 66.83 & 28.00 & 60.10 & \textbf{27.70} & 24.00 & 11.84 & \textbf{81.62} & \underline{22.79} & \textbf{43.03} \\
        \textbf{\ocr} & 50.81 & 54.38 & 36.06 & \textbf{46.88} & \textbf{68.22} & \underline{29.50} & 57.06 & 27.09 & 24.30 & \textbf{12.27} & 80.82 & 19.47 & 42.24  \\
        \textbf{\matht} & 48.87 & \underline{55.93} & 36.41 & \underline{44.24} & 67.23 & \textbf{29.90} & \underline{60.30} & \underline{27.69} & \textbf{26.50} & 11.97 & \underline{81.04} & 22.09 & \underline{42.68}  \\
        \textbf{\hall} & \underline{55.31} & 53.44 & 35.18 & 38.18 & 67.30 & 25.70 & \textbf{60.71} & 26.30 & 23.10 & 11.74 & 80.46 & \textbf{24.87} & 41.86  \\

\bottomrule
\end{tabular}
}
\label{table:specific_each_benchmark}
\end{table*}

\begin{table*}[h!]
\centering
\caption{Performance of specific knowledge-oriented constrains on fine-grained knowledge types with LLaVA-v1.5 {\small(7B)}.}
\vspace{-6pt}
\label{tab:specific_knowledge_type_appendix}


\renewcommand{\arraystretch}{1.2} 
\resizebox{\textwidth}{!}{%
\begin{tabular}{l|cc|cc|cc|cc|cc|cc|cc|cc|cc|cc}
\toprule
\multirow{4}{*}{\textbf{Method}} & \multicolumn{10}{c|}{\textbf{News}} & \multicolumn{10}{c}{\textbf{Entity}} \\
\cmidrule{2-11} \cmidrule{12-21}

& \multicolumn{2}{c|}{\textbf{Avg}} & \multicolumn{2}{c|}{\textbf{PO}} & \multicolumn{2}{c|}{\textbf{SP}} & \multicolumn{2}{c|}{\textbf{BU}} & \multicolumn{2}{c|}{\textbf{HE}} & \multicolumn{2}{c|}{\textbf{Avg}} & \multicolumn{2}{c|}{\textbf{CE}} & \multicolumn{2}{c|}{\textbf{FI}} & \multicolumn{2}{c|}{\textbf{AL}} & \multicolumn{2}{c}{\textbf{WR}} \\
\cmidrule{2-3} \cmidrule{4-5} \cmidrule{6-7} \cmidrule{8-9} \cmidrule{10-11} \cmidrule{12-13} \cmidrule{14-15} \cmidrule{16-17} \cmidrule{18-19} \cmidrule{20-21}

& \textbf{CEM} \daugshifted & \textbf{F1} \daugshifted
& \textbf{CEM} \daugshifted & \textbf{F1} \daugshifted
& \textbf{CEM} \daugshifted & \textbf{F1} \daugshifted
& \textbf{CEM} \daugshifted & \textbf{F1} \daugshifted
& \textbf{CEM} \daugshifted & \textbf{F1} \daugshifted
& \textbf{CEM} \daugshifted & \textbf{F1} \daugshifted
& \textbf{CEM} \daugshifted & \textbf{F1} \daugshifted
& \textbf{CEM} \daugshifted & \textbf{F1} \daugshifted
& \textbf{CEM} \daugshifted & \textbf{F1} \daugshifted
& \textbf{CEM} \daugshifted & \textbf{F1} \daugshifted \\
\midrule

\rowcolor{backblue!80}
       \textbf{\method} & \textbf{34.74} & \textbf{42.96} & 23.83 & \textbf{32.31} & \textbf{46.19} & \underline{50.38} & 34.69 & 45.74 & \underline{33.20} & \underline{45.23} & \textbf{26.17} & \textbf{39.39} & \underline{27.79} & \textbf{42.61} & \textbf{26.93} & \textbf{34.05} & \textbf{16.52} & \textbf{29.54} & \underline{28.81} & \textbf{43.05} \\

\midrule

\textbf{\mme} & 34.05 & 41.53 & 23.92 & 31.46 & 43.17 & 47.28 & 34.32 & \textbf{46.12} & \textbf{35.63} & \textbf{45.38} & 24.48 & 37.15 & 27.24 & 40.96 & 22.61 & 30.43 & \underline{15.07} & \underline{27.72} & \textbf{30.51} & 42.16 \\
\textbf{\ocr} & \underline{34.46} & 41.66 & \textbf{24.29} & 31.69 & 43.53 & 48.34 & \textbf{36.35} & \underline{46.09} & \underline{33.20} & 44.35 & 25.01 & 37.65 & 27.24 & 41.17 & 24.09 & 31.60 & 14.78 & 27.16 & \textbf{30.51} & \underline{42.17} \\
\textbf{\matht} & 33.71 & 41.72 & 22.27 & 30.39 & \underline{45.95} & \textbf{50.88} & 33.03 & 43.38 & 30.77 & 43.55 & \underline{26.06} & \underline{38.82} & \textbf{28.15} & \underline{42.46} & \underline{25.80} & \underline{32.97} & \underline{15.07} & 27.37 & \textbf{30.51} & 42.11 \\
\textbf{\hall} & 34.23 & \underline{41.74} & \underline{24.11} & \underline{32.09} & 43.05 & 46.98 & \underline{35.06} & 44.92 & 32.39 & 43.53 & 25.21 & 38.05 & 27.54 & 41.68 & 24.66 & 32.34 & 14.78 & 26.86 & \underline{28.81} & 40.13 \\
\bottomrule
\end{tabular}%
}
\end{table*}

\textbf{Obs 3 in \autoref{appendix:specific}: Specific knowledge-oriented constraints also achieve excellent adaptation performance across a wide spectrum of fine-grained knowledge.} In Table~\ref{tab:specific_knowledge_type_appendix}, \method with specific knowledge-oriented constraints maintains strong adaptation performance across various News and Entity knowledge types, with negligible performance degradation.

\subsection{More Results On Ablation Experiments}

Regarding the experiment in \autoref{sec:ablation}, we have supplemented the experiments in \autoref{appendix:ablation_1} and \autoref{appendix:ablation_2}.

\subsubsection{Rank Ablation Experiments}\label{appendix:ablation_1}

\begin{table*}[h]
\caption{Performance comparison of different ranks with LLaVA-v1.5 {\small(7B)}.}
    \label{tab:rank_main_appendix}
  \vspace{-6pt}
  \centering
  \renewcommand{\arraystretch}{1.2} 
  \resizebox{\textwidth}{!}{%
    \begin{tabular}{l|cc|ccccccc|cccc}
      \toprule
      \multirow{2.5}{*}{\textbf{Method}}
      & \multicolumn{2}{c|}{\textbf{\dataset}}
      & \multirow{2.5}{*}{\textbf{COM} \daugshifted}
      & \multirow{2.5}{*}{\textbf{OCR} \daugshifted}
      & \multirow{2.5}{*}{\textbf{M-DIS} \daugshifted}
      & \multirow{2.5}{*}{\textbf{INS} \daugshifted}
      & \multirow{2.5}{*}{\textbf{M-IDU} \daugshifted}
      & \multirow{2.5}{*}{\textbf{MAT} \daugshifted}
      & \multirow{2.5}{*}{\textbf{HAL} \daugshifted}
      & \multirow{2.5}{*}{\textbf{K.A} \daugshifted}
      & \multirow{2.5}{*}{\textbf{K.R} \daugshifted}
      & \multirow{2.5}{*}{\textbf{Avg} \daugshifted}
      & \multirow{2.5}{*}{\textbf{HARS} \daugshifted} \\
      \cmidrule{2-3} 
      & \textbf{CEM} \daugshifted & \textbf{F1}\daugshifted  & & & & & & & & & & & \\

      \midrule
\rowcolor{backblue!80}

\textbf{\method(rank=64)} & 24.00 & 33.07 & 45.35 & 29.46 & 45.02 & \underline{44.07} & \underline{19.62} & 18.08 & 44.48 & 28.54 & 35.70 & 32.12 & 75.73 \\
\textbf{\method(rank=128)} & \underline{30.72} & 40.55 & 49.97 & 36.05 & 47.07 & 34.87 & 10.00 & 17.46 & 50.30 & 35.64 & 37.21 & 36.42 & 79.83 \\

       \textbf{\method(rank=235)} & 30.65 & \underline{41.26} & \underline{52.41} & \textbf{40.98} & \underline{48.68} & 38.54 & 16.58 & \textbf{18.59} & \textbf{51.75} & \underline{35.96} & \underline{40.00} & \underline{37.98} & \underline{82.81} \\

\textbf{\method(rank=256)} & \textbf{31.05} & \textbf{41.32} & \textbf{52.48} & \underline{39.96} & \textbf{48.96} & \textbf{60.02} & \textbf{23.18} & \underline{18.09} & \underline{51.50} & \textbf{36.19} & \textbf{42.10} & \textbf{39.14} & \textbf{84.93} \\

      \bottomrule
    \end{tabular}%
  }
\end{table*}

\textbf{Obs 1 in \autoref{appendix:ablation_1}: Increasing the number of trainable parameters enables \method to achieve stronger performance.} In Table~\ref{tab:rank_main_appendix}, \method's performance in both knowledge adaptation and knowledge retention exhibits a consistent upward trend as the rank and number of trainable parameters increase. This trend is particularly significant on the INS and M-IDU dimensions, which indicates \method's potential to achieve even stronger performance with larger parameter.

\begin{table*}[h!]
\caption{Performance of comparison across different ranks on fine-grained knowledge retention evaluations with LLaVA-v1.5 {\small(7B)}.}
\vspace{-6pt}

\centering
\resizebox{1\linewidth}{!}{
\begin{tabular}{l|c c|c c|c c|c|c|c c|c c|c}
\toprule
\multirow{2.5}{*}{\textbf{Method}} & \multicolumn{2}{c|}{\textbf{COM}} & \multicolumn{2}{c|}{\textbf{OCR}} & \multicolumn{2}{c|}{\textbf{M-DIS}} & \multicolumn{1}{c|}{\textbf{INS}} & \multicolumn{1}{c|}{\textbf{M-IDU}} & \multicolumn{2}{c|}{\textbf{MAT}} & \multicolumn{2}{c|}{\textbf{HAL}} & \multirow{2.5}{*}{\textbf{Avg}} \\
\cmidrule{2-13}
& \textbf{MME} \daugshifted & \textbf{MM}\raisebox{1ex}{\tiny\textbf{B}} \daugshifted & \textbf{SEED}\raisebox{1ex}{\tiny\textbf{B2P}} \daugshifted & \textbf{OCR}\raisebox{1ex}{\tiny\textbf{VQA}} \daugshifted & \textbf{SQA} \daugshifted & \textbf{MMMU}\raisebox{1ex}{\tiny\textbf{T}} \daugshifted &\textbf{MIA}\raisebox{1ex}{\tiny\textbf{B}} \daugshifted & \textbf{MMDU} \daugshifted & \textbf{Math}\raisebox{1ex}{\tiny\textbf{T}} \daugshifted & \textbf{Math}\raisebox{1ex}{\tiny\textbf{I}} \daugshifted & \textbf{POPE} \daugshifted & \textbf{Hall}\raisebox{1ex}{\tiny\textbf{B}} \daugshifted \\
\midrule

      \textbf{\method (rank=64)}      &43.63&47.08&33.55&25.36&66.34&23.70&\underline{44.07}&\underline{19.62}&\textbf{25.20}&10.95&74.22&14.73&35.70   \\

      \textbf{\method (rank=128)}      &47.96&51.98&36.32&35.77&67.44&26.70&34.87&10.00&23.90&11.02&79.63&20.97&37.21   \\

      \rowcolor{backblue!80}
      \textbf{\method (rank=235)}      &\underline{49.84}&\textbf{54.98}&\textbf{37.73}&\textbf{44.24}&\underline{68.06}&\underline{29.30}&38.54&16.58&\underline{25.10}&\textbf{12.09}&\textbf{80.99}&\textbf{22.51}&\underline{40.00}   \\

      \textbf{\method (rank=256)}      &\textbf{50.06}&\underline{54.90}&\underline{36.89}&\underline{43.03}&\textbf{68.51}&\textbf{29.40}&\textbf{60.02}&\textbf{23.18}&24.70&\underline{11.48}&\underline{80.77}&\underline{22.23}&\textbf{42.10}   \\

\bottomrule
\end{tabular}

}
\label{table:rank_each_benchmark}
\end{table*}

\textbf{Obs 2 in \autoref{appendix:ablation_1}: Larger trainable parameter scales enhance \method's knowledge retention performance.} In Table~\ref{table:rank_each_benchmark}, \method (rank=256) achieves near-comprehensive superiority across 12 benchmarks and surpasses \method (rank=235) by $2.10$ in overall performance. This underscores that a larger trainable parameter scale activates stronger knowledge retention in \method.

\begin{table*}[ht]
\centering
\caption{Performance comparison across different ranks on fine-grained knowledge types with LLaVA-v1.5 {\small(7B)}.}
\label{tab:rank_knowledge}

\vspace{-6pt}

\renewcommand{\arraystretch}{1.2} 
\resizebox{\textwidth}{!}{%
\begin{tabular}{l|cc|cc|cc|cc|cc|cc|cc|cc|cc|cc}
\toprule
\multirow{3}{*}{\textbf{Method}} & \multicolumn{10}{c|}{\textbf{News}} & \multicolumn{10}{c}{\textbf{Entity}} \\
\cmidrule(lr){2-11} \cmidrule(lr){12-21}

& \multicolumn{2}{c|}{\textbf{Avg}} & \multicolumn{2}{c|}{\textbf{PO}} & \multicolumn{2}{c|}{\textbf{SP}} & \multicolumn{2}{c|}{\textbf{BU}} & \multicolumn{2}{c|}{\textbf{HE}} & \multicolumn{2}{c|}{\textbf{Avg}} & \multicolumn{2}{c|}{\textbf{CE}} & \multicolumn{2}{c|}{\textbf{FI}} & \multicolumn{2}{c|}{\textbf{AL}} & \multicolumn{2}{c}{\textbf{WR}} \\
\cmidrule(lr){2-3} \cmidrule(lr){4-5} \cmidrule(lr){6-7} \cmidrule(lr){8-9} \cmidrule(lr){10-11} \cmidrule(lr){12-13} \cmidrule(lr){14-15} \cmidrule(lr){16-17} \cmidrule(lr){18-19} \cmidrule(lr){20-21}

& \textbf{CEM} \daugshifted & \textbf{F1} \daugshifted
& \textbf{CEM} \daugshifted & \textbf{F1} \daugshifted
& \textbf{CEM} \daugshifted & \textbf{F1} \daugshifted
& \textbf{CEM} \daugshifted & \textbf{F1} \daugshifted
& \textbf{CEM} \daugshifted & \textbf{F1} \daugshifted
& \textbf{CEM} \daugshifted & \textbf{F1} \daugshifted
& \textbf{CEM} \daugshifted & \textbf{F1} \daugshifted
& \textbf{CEM} \daugshifted & \textbf{F1} \daugshifted
& \textbf{CEM} \daugshifted & \textbf{F1} \daugshifted
& \textbf{CEM} \daugshifted & \textbf{F1} \daugshifted \\
\midrule

\textbf{\method (rank=64)} & 28.31 & 34.84 & 20.44 & 27.66 & 36.64 & 41.11 & 28.60 & 38.13 & 26.72 & 35.77 & 19.27 & 31.11 & 21.24 & 35.25 & 18.98 & 25.33 & 11.01 & 23.14 & 22.03 & 33.44 \\
\textbf{\method (rank=128)} & 34.70 & 42.07 & \textbf{24.20} & \underline{31.56} & 44.50 & 49.17 & \textbf{36.72} & \textbf{47.68} & \textbf{34.82} & \underline{44.39} & \underline{26.35} & 38.89 & \textbf{28.81} & \textbf{43.19} & 23.86 & 30.22 & \textbf{17.97} & \underline{28.77} & \textbf{35.59} & \textbf{44.86} \\

\rowcolor{backblue!80}
\textbf{\method (rank=235)} & \underline{34.74} & \underline{42.96} & 23.83 & \textbf{32.31} & \textbf{46.19} & \textbf{50.38} & 34.69 & 45.74 & \underline{33.20} & \textbf{45.23} & 26.17 & \underline{39.39} & 27.79 & 42.61 & \underline{26.93} & \underline{34.05} & \underline{16.52} & \textbf{29.54} & \underline{28.81} & \underline{43.05} \\

\textbf{\method (rank=256)} & \textbf{35.17} & \textbf{42.98} & \underline{23.92} & 31.24 & \underline{45.83} & \underline{50.35} & \underline{35.98} & \underline{47.11} & 32.79 & 43.80 & \textbf{26.55} & \textbf{39.49} & \underline{28.46} & \underline{42.74} & \textbf{27.16} & \textbf{34.52} & 15.65 & 26.81 & 27.12 & 39.92 \\
\bottomrule
\end{tabular}%
}
\end{table*}

\textbf{Obs 3 in \autoref{appendix:ablation_1}: Larger trainable parameters improve \method's knowledge adaptation performance on News and Entity types.} In Table~\ref{tab:rank_knowledge}, \method (rank=256) achieves robust and consistent performance across a broader range of fine-grained knowledge types, demonstrating \method's potential for superior performance with an increased number of trainable parameters.

\subsubsection{Setting Ablation Experiments}\label{appendix:ablation_2}

\begin{table*}[h]
\caption{Performance comparison of setting ablation with LLaVA-v1.5 {\small(7B)}.}
\label{tab:setting_main_result}
  \vspace{-6pt}
  \centering
  \renewcommand{\arraystretch}{1.2} 
  \resizebox{\textwidth}{!}{%
    \begin{tabular}{l|cc|ccccccc|cccc}
      \toprule
      \multirow{2.5}{*}{\textbf{Method}}
      & \multicolumn{2}{c|}{\textbf{\dataset}}
      & \multirow{2.5}{*}{\textbf{COM} \daugshifted}
      & \multirow{2.5}{*}{\textbf{OCR} \daugshifted}
      & \multirow{2.5}{*}{\textbf{M-DIS} \daugshifted}
      & \multirow{2.5}{*}{\textbf{INS} \daugshifted}
      & \multirow{2.5}{*}{\textbf{M-IDU} \daugshifted}
      & \multirow{2.5}{*}{\textbf{MAT} \daugshifted}
      & \multirow{2.5}{*}{\textbf{HAL} \daugshifted}
      & \multirow{2.5}{*}{\textbf{K.A} \daugshifted}
      & \multirow{2.5}{*}{\textbf{K.R} \daugshifted}
      & \multirow{2.5}{*}{\textbf{Avg} \daugshifted}
      & \multirow{2.5}{*}{\textbf{HARS} \daugshifted} \\
      \cmidrule{2-3} 
      & \textbf{CEM} \daugshifted & \textbf{F1}\daugshifted  & & & & & & & & & & & \\

      \midrule

\rowcolor{backblue!80}
       \textbf{\method} & 30.65 & 41.26 & \underline{52.41} & \textbf{40.98} & \textbf{48.68} & \textbf{38.54} & \textbf{16.58} & 18.59 & \textbf{51.75} & 35.96 & \underline{40.00} & \textbf{37.98} & \textbf{82.81} \\

      \midrule

W/o Augmentation & 10.83 & 18.31 & \textbf{59.96} & \underline{40.42} & 47.13 & 32.53 & 16.00 & \textbf{19.71} & 49.50 & 14.57 & \textbf{40.16} & 27.37 & 64.14 \\
W/o Constraint & \textbf{33.93} & \textbf{43.71} & 46.39 & 32.38 & 46.31 & 32.70 & 15.38 & \underline{19.12} & 46.47 & \textbf{38.82} & 35.78 & 37.30 & 79.04 \\
W/o Frozen Matrix $A$ & \underline{31.97} & \underline{41.72} & 50.73 & 39.56 & \underline{48.37} & \underline{35.30} & \underline{16.44} & 19.07 & \underline{49.91} & \underline{36.85} & 38.92 & \underline{37.88} & \underline{81.96} \\
      
      \bottomrule
    \end{tabular}%
  }
\end{table*}

\begin{table*}[h!]
\caption{Performance comparison  of setting ablation in knowledge retention with LLaVA-v1.5 {\small(7B)}.}

\vspace{-6pt}
\centering
\resizebox{1\linewidth}{!}{
\begin{tabular}{l|c c|c c|c c|c|c|c c|c c|c}
\toprule
\multirow{2.5}{*}{\textbf{Method}} & \multicolumn{2}{c|}{\textbf{COM}} & \multicolumn{2}{c|}{\textbf{OCR}} & \multicolumn{2}{c|}{\textbf{M-DIS}} & \multicolumn{1}{c|}{\textbf{INS}} & \multicolumn{1}{c|}{\textbf{M-IDU}} & \multicolumn{2}{c|}{\textbf{MAT}} & \multicolumn{2}{c|}{\textbf{HAL}} & \multirow{2.5}{*}{\textbf{Avg}} \\
\cmidrule{2-13}
& \textbf{MME} \daugshifted & \textbf{MM}\raisebox{1ex}{\tiny\textbf{B}} \daugshifted & \textbf{SEED}\raisebox{1ex}{\tiny\textbf{B2P}} \daugshifted & \textbf{OCR}\raisebox{1ex}{\tiny\textbf{VQA}} \daugshifted & \textbf{SQA} \daugshifted & \textbf{MMMU}\raisebox{1ex}{\tiny\textbf{T}} \daugshifted &\textbf{MIA}\raisebox{1ex}{\tiny\textbf{B}} \daugshifted & \textbf{MMDU} \daugshifted & \textbf{Math}\raisebox{1ex}{\tiny\textbf{T}} \daugshifted & \textbf{Math}\raisebox{1ex}{\tiny\textbf{I}} \daugshifted & \textbf{POPE} \daugshifted & \textbf{Hall}\raisebox{1ex}{\tiny\textbf{B}} \daugshifted \\
\midrule

\rowcolor{backblue!80}
\textbf{\method} 
& \underline{49.84} & \underline{54.98} & \textbf{37.73} & \textbf{44.24} & \underline{68.06} & \textbf{29.30} & \textbf{38.54} & \textbf{16.58} & 25.10 & \textbf{12.09} & \underline{80.99} & \textbf{22.51} & \textbf{51.75} \\

\midrule

 W/o Augmentation & \textbf{58.75} & \textbf{61.17} & \underline{36.80} & \underline{44.04} & \textbf{68.15} & 26.10 & 32.53 & 16.00 & \textbf{28.00} & 11.41 & \textbf{81.29} & 17.71 & \underline{40.16}      \\

        W/o Constraint & 40.55 & 52.23 & 31.75 & 33.01 & 65.81 & 26.80 & 32.70 & 15.38 & 26.50 & \underline{11.74} & 79.16 & 13.77 & 35.78         \\
        W/o Frozen Matrix $A$ & 47.24 & 54.21 & 36.01 & 43.10 & 67.63 & \underline{29.10} & \underline{35.30} & \underline{16.44} & \underline{26.70} & 11.45 & 80.84 & \underline{18.98} & 38.92  \\

\bottomrule
\end{tabular}
}
\label{table:benchmark_ablation}
\end{table*}

\textbf{Obs 1 in \autoref{appendix:ablation_2}: Modifying \method's design leads to a degradation in overall knowledge retention performance.} In Table~\ref{table:benchmark_ablation}, the ablated versions W/o Augmentation, W/o Constraint, and W/o Frozen Matrix $A$ exhibit overall performance degradations of $11.59$, $15.97$, and $12.83$ respectively compared to \method. This significant degradation underscores the high efficacy of \method's design.

\begin{table*}[h!]
\centering
\caption{Performance comparison of setting ablation on fine-grained knowledge types with LLaVA-v1.5 {\small(7B)}.}
\label{tab:ablation_knowledge}

\vspace{-6pt}
\renewcommand{\arraystretch}{1.2} 
\resizebox{\textwidth}{!}{%
\begin{tabular}{l|cc|cc|cc|cc|cc|cc|cc|cc|cc|cc}
\toprule
\multirow{4}{*}{\textbf{Method}} & \multicolumn{10}{c|}{\textbf{News}} & \multicolumn{10}{c}{\textbf{Entity}} \\
\cmidrule{2-11} \cmidrule{12-21}

& \multicolumn{2}{c|}{\textbf{Avg}} & \multicolumn{2}{c|}{\textbf{PO}} & \multicolumn{2}{c|}{\textbf{SP}} & \multicolumn{2}{c|}{\textbf{BU}} & \multicolumn{2}{c|}{\textbf{HE}} & \multicolumn{2}{c|}{\textbf{Avg}} & \multicolumn{2}{c|}{\textbf{CE}} & \multicolumn{2}{c|}{\textbf{FI}} & \multicolumn{2}{c|}{\textbf{AL}} & \multicolumn{2}{c}{\textbf{WR}} \\
\cmidrule{2-3} \cmidrule{4-5} \cmidrule{6-7} \cmidrule{8-9} \cmidrule{10-11} \cmidrule{12-13} \cmidrule{14-15} \cmidrule{16-17} \cmidrule{18-19} \cmidrule{20-21}

& \textbf{CEM} \daugshifted & \textbf{F1} \daugshifted
& \textbf{CEM} \daugshifted & \textbf{F1} \daugshifted
& \textbf{CEM} \daugshifted & \textbf{F1} \daugshifted
& \textbf{CEM} \daugshifted & \textbf{F1} \daugshifted
& \textbf{CEM} \daugshifted & \textbf{F1} \daugshifted
& \textbf{CEM} \daugshifted & \textbf{F1} \daugshifted
& \textbf{CEM} \daugshifted & \textbf{F1} \daugshifted
& \textbf{CEM} \daugshifted & \textbf{F1} \daugshifted
& \textbf{CEM} \daugshifted & \textbf{F1} \daugshifted
& \textbf{CEM} \daugshifted & \textbf{F1} \daugshifted \\
\midrule

\rowcolor{backblue!80}
\textbf{\method}  & 34.74 & 42.96 & 23.83 & \underline{32.31} & \underline{46.19} & 50.38 & 34.69 & 45.74 & 33.20 & 45.23 & 26.17 & 39.39 & 27.79 & 42.61 & 26.93 & 34.05 & 16.52 & 29.54 & \textbf{28.81} & \textbf{43.05} \\
W/o Augmentation & 14.04 & 20.22 & 8.25 & 14.06 & 15.96 & 20.08 & 14.39 & 23.13 & 14.57 & 25.69 & 7.30 & 16.21 & 8.08 & 20.13 & 8.41 & 14.15 & 3.77 & 6.56 & \underline{13.56} & 22.27 \\
W/o Constraint & \textbf{38.45} & \textbf{45.16} & \textbf{25.57} & \textbf{32.56} & \textbf{46.43} & \textbf{50.66} & \textbf{41.33} & \textbf{51.22} & \textbf{36.84} & \underline{45.78} & \textbf{28.97} & \textbf{42.12} & \textbf{29.67} & \textbf{44.18} & \textbf{30.45} & \textbf{37.75} & \textbf{20.58} & \textbf{33.59} & \textbf{28.81} & \underline{40.02} \\
W/o Frozen Matrix $A$ & \underline{36.49} & \underline{43.42} & \underline{25.11} & 31.70 & \textbf{46.43} & \underline{50.44} & \underline{37.82} & \underline{48.20} & \underline{36.44} & \textbf{46.44} & \underline{27.01} & \underline{39.85} & \underline{28.05} & \underline{42.88} & \underline{27.95} & \underline{34.57} & \underline{19.13} & \underline{30.95} & \textbf{28.81} & 39.86 \\
\bottomrule
\end{tabular}%
}
\end{table*}

\textbf{Obs 2 in \autoref{appendix:ablation_2}: W/o Constraint yields superior knowledge adaptation performance across a wide spectrum of fine-grained knowledge.} In Table~\ref{tab:ablation_knowledge}, W/o Constraint achieves superior knowledge adaptation performance on fine-grained News and Entity types. These gains stem from \aug's ability to perform profound and structured augmentation.

\subsection{More Results On Comparison With General Augmentation methods}\label{appendix:Augmentation_Comparison}

\begin{table*}[ht]
\caption{Performance comparison of different augmentation methods with LLaVA-v1.5 {\small(7B)}.}
  \label{tab:augmentation_overall}
  \vspace{-6pt}
  \centering
  \renewcommand{\arraystretch}{1.2} 
  \resizebox{\textwidth}{!}{%
    \begin{tabular}{l|cc|ccccccc|cccc}
      \toprule
      \multirow{2.5}{*}{\textbf{Method}}
      & \multicolumn{2}{c|}{\textbf{\dataset}}
      & \multirow{2.5}{*}{\textbf{COM} \daugshifted}
      & \multirow{2.5}{*}{\textbf{OCR} \daugshifted}
      & \multirow{2.5}{*}{\textbf{M-DIS} \daugshifted}
      & \multirow{2.5}{*}{\textbf{INS} \daugshifted}
      & \multirow{2.5}{*}{\textbf{M-IDU} \daugshifted}
      & \multirow{2.5}{*}{\textbf{MAT} \daugshifted}
      & \multirow{2.5}{*}{\textbf{HAL} \daugshifted}
      & \multirow{2.5}{*}{\textbf{K.A} \daugshifted}
      & \multirow{2.5}{*}{\textbf{K.R} \daugshifted}
      & \multirow{2.5}{*}{\textbf{Avg} \daugshifted}
      & \multirow{2.5}{*}{\textbf{HARS} \daugshifted} \\
      \cmidrule{2-3} 
      & \textbf{CEM} \daugshifted & \textbf{F1}\daugshifted  & & & & & & & & & & & \\

      \midrule
\rowcolor{backblue!80}
\textbf{KORE-AUGMENTATION} & \textbf{33.93} & \textbf{43.71} & 46.39 & \textbf{32.38} & \textbf{46.31} & 32.70 & 15.38 & \underline{19.12} & \textbf{46.47} & \textbf{38.82} & \textbf{35.78} & \textbf{37.30} & \textbf{79.04} \\

      \midrule

\rowcolor{gray!10}

\multicolumn{14}{c}{\fontsize{10}{12}\selectfont \textit{\textbf{Augmentation for Text}}} \\

      \midrule
Knowledge-Agnostic & 12.10 & 19.10 & \textbf{53.50} & 25.38 & 39.79 & \textbf{44.37} & \textbf{22.41} & 18.47 & \underline{43.72} & 15.60 & \underline{35.71} & 25.65 & 63.54 \\
Knowledge-Aware  & \underline{16.94} & \underline{23.64} & 48.85 & \underline{28.10} & 40.28 & 43.59 & 20.72 & 18.14 & 41.65 & \underline{20.29} & 34.86 & \underline{27.58} & \underline{69.43} \\

      \midrule

\rowcolor{gray!10}

\multicolumn{14}{c}{\fontsize{10}{12}\selectfont \textit{\textbf{Augmentation for Images}}} \\

      \midrule
Knowledge-Agnostic & 15.46 & 21.20 & \underline{51.42} & 13.22 & 36.94 & \underline{43.65} & \underline{21.60} & 17.09 & 41.27 & 18.33 & 32.09 & 25.21 & 64.71 \\
Knowledge-Aware  & 14.68 & 21.97 & 51.40 & 24.57 & \underline{40.54} & 40.84 & 18.53 & \textbf{19.23} & 38.70 & 18.33 & 34.02 & 26.17 & 66.48 \\

      \bottomrule
    \end{tabular}%
  }
\end{table*}

\begin{table*}[ht]
\caption{Performance comparison  of different augmentation methods in knowledge retention with LLaVA-v1.5 {\small(7B)}.}

\vspace{-6pt}

\centering

\resizebox{1\linewidth}{!}{
\begin{tabular}{l|c c|c c|c c|c|c|c c|c c|c}
\toprule
\multirow{2.5}{*}{\textbf{Method}} & \multicolumn{2}{c|}{\textbf{COM}} & \multicolumn{2}{c|}{\textbf{OCR}} & \multicolumn{2}{c|}{\textbf{M-DIS}} & \multicolumn{1}{c|}{\textbf{INS}} & \multicolumn{1}{c|}{\textbf{M-IDU}} & \multicolumn{2}{c|}{\textbf{MAT}} & \multicolumn{2}{c|}{\textbf{HAL}} & \multirow{2.5}{*}{\textbf{Avg}} \\
\cmidrule{2-13}
& \textbf{MME} \daugshifted & \textbf{MM}\raisebox{1ex}{\tiny\textbf{B}} \daugshifted & \textbf{SEED}\raisebox{1ex}{\tiny\textbf{B2P}} \daugshifted & \textbf{OCR}\raisebox{1ex}{\tiny\textbf{VQA}} \daugshifted & \textbf{SQA} \daugshifted & \textbf{MMMU}\raisebox{1ex}{\tiny\textbf{T}} \daugshifted &\textbf{MIA}\raisebox{1ex}{\tiny\textbf{B}} \daugshifted & \textbf{MMDU} \daugshifted & \textbf{Math}\raisebox{1ex}{\tiny\textbf{T}} \daugshifted & \textbf{Math}\raisebox{1ex}{\tiny\textbf{I}} \daugshifted & \textbf{POPE} \daugshifted & \textbf{Hall}\raisebox{1ex}{\tiny\textbf{B}} \daugshifted \\
\midrule

\rowcolor{backblue!80}
         \textbf{\aug} 
&40.55&52.23&\textbf{31.75}&\textbf{33.01}&\underline{65.81}&\textbf{26.80}&32.70&15.38&\textbf{26.50}&11.74&\textbf{79.16}&\textbf{13.77}&\textbf{46.47} \\

\midrule
    \rowcolor{gray!10}
        \multicolumn{14}{c}{\fontsize{10}{12}\selectfont \textit{\textbf{Augmentation for Text}}} \\

        Knowledge-Agnostic &\textbf{51.67}&\textbf{55.33}&25.99&24.77&64.38&\underline{15.20}&\textbf{44.37}&\textbf{22.41}&\underline{25.20}&11.74&\underline{79.04}&8.40&\underline{35.71}  \\

               Knowledge-Aware &50.02&47.68&24.95&\underline{31.25}&65.75&14.80&43.59&20.72&24.20&\underline{12.07}&74.05&\underline{9.24}&34.86  \\

\midrule
    \rowcolor{gray!10}
        \multicolumn{14}{c}{\fontsize{10}{12}\selectfont \textit{\textbf{Augmentation for Images}}} \\

        Knowledge-Agnostic &50.43&\underline{52.41}&11.86&14.58&64.18&9.70&\underline{43.65}&\underline{21.60}&22.60&11.58&73.95&8.58&32.09  \\

               Knowledge-Aware &\underline{51.35}&51.46&\underline{27.23}&21.91&\textbf{66.29}&14.80&40.84&18.53&21.20&\textbf{17.26}&69.71&7.68&34.02  \\

\bottomrule
\end{tabular}
}
\label{table:benchmark_com_aug}
\end{table*}

\begin{table*}[h!]
\centering
\caption{Performance comparison of different augmentation methods on fine-grained knowledge types with LLaVA-v1.5 {\small(7B)}.}
\label{table:knowledge_com_aug}
\vspace{-6pt}

\renewcommand{\arraystretch}{1.2} 
\resizebox{\textwidth}{!}{%
\begin{tabular}{l|cc|cc|cc|cc|cc|cc|cc|cc|cc|cc}
\toprule
\multirow{4}{*}{\textbf{Method}} & \multicolumn{10}{c|}{\textbf{News}} & \multicolumn{10}{c}{\textbf{Entity}} \\
\cmidrule{2-11} \cmidrule{12-21}

& \multicolumn{2}{c|}{\textbf{Avg}} & \multicolumn{2}{c|}{\textbf{PO}} & \multicolumn{2}{c|}{\textbf{SP}} & \multicolumn{2}{c|}{\textbf{BU}} & \multicolumn{2}{c|}{\textbf{HE}} & \multicolumn{2}{c|}{\textbf{Avg}} & \multicolumn{2}{c|}{\textbf{CE}} & \multicolumn{2}{c|}{\textbf{FI}} & \multicolumn{2}{c|}{\textbf{AL}} & \multicolumn{2}{c}{\textbf{WR}} \\
\cmidrule{2-3} \cmidrule{4-5} \cmidrule{6-7} \cmidrule{8-9} \cmidrule{10-11} \cmidrule{12-13} \cmidrule{14-15} \cmidrule{16-17} \cmidrule{18-19} \cmidrule{20-21}

& \textbf{CEM} \daugshifted & \textbf{F1} \daugshifted
& \textbf{CEM} \daugshifted & \textbf{F1} \daugshifted
& \textbf{CEM} \daugshifted & \textbf{F1} \daugshifted
& \textbf{CEM} \daugshifted & \textbf{F1} \daugshifted
& \textbf{CEM} \daugshifted & \textbf{F1} \daugshifted
& \textbf{CEM} \daugshifted & \textbf{F1} \daugshifted
& \textbf{CEM} \daugshifted & \textbf{F1} \daugshifted
& \textbf{CEM} \daugshifted & \textbf{F1} \daugshifted
& \textbf{CEM} \daugshifted & \textbf{F1} \daugshifted
& \textbf{CEM} \daugshifted & \textbf{F1} \daugshifted \\
\midrule

\rowcolor{backblue!80}
\textbf{\aug}  & \textbf{38.45} & \textbf{45.16} & \textbf{25.57} & \textbf{32.56} & \textbf{46.43} & \textbf{50.66} & \textbf{41.33} & \textbf{51.22} & \textbf{36.84} & \textbf{45.78} & \textbf{28.97} & \textbf{42.12} & \textbf{29.67} & \textbf{44.18} & \textbf{30.45} & \textbf{37.75} & \textbf{20.58} & \textbf{33.59} & \textbf{28.81} & \textbf{40.02} \\

\rowcolor{gray!10}
\multicolumn{21}{c}{\fontsize{10}{12}\selectfont \textit{\textbf{Augmentation for Text}}}\\
Knowledge-Agnostic & 14.59 & 20.11 & 8.52 & 14.84 & 17.05 & 21.34 & 17.16 & 24.56 & 14.57 & 23.34 & 9.37 & 17.99 & 10.52 & 22.06 & 6.59 & 10.74 & 8.12 & 15.00 & 13.56 & 21.25 \\
Knowledge-Aware & \underline{20.19} & \underline{24.99} & \underline{11.37} & \underline{16.28} & \underline{24.55} & \underline{28.48} & \underline{21.96} & \underline{29.00} & 19.03 & \underline{28.94} & \underline{13.37} & \underline{22.17} & \underline{13.92} & \underline{26.33} & 12.95 & 18.16 & 9.57 & 12.99 & 13.56 & 18.90 \\

\rowcolor{gray!10}
\multicolumn{21}{c}{\fontsize{10}{12}\selectfont \textit{\textbf{Augmentation for Images}}}\\
Knowledge-Agnostic & 18.38 & 22.42 & 10.72 & 14.88 & 22.97 & 26.92 & 20.11 & 26.60 & \underline{19.84} & 27.28 & 12.26 & 19.87 & 12.35 & 23.27 & 13.07 & 16.59 & \underline{10.43} & \underline{16.13} & \underline{15.25} & 14.83 \\
Knowledge-Aware & 17.15 & 23.01 & 9.99 & 14.93 & 19.35 & 24.35 & 18.08 & 25.79 & 16.19 & 27.05 & 11.97 & 20.84 & 12.86 & 24.29 & \underline{13.86} & \underline{19.59} & 7.25 & 11.47 & \underline{15.25} & \underline{21.78} \\

\bottomrule
\end{tabular}%
}
\end{table*}

\textbf{Obs 1 in \autoref{appendix:Augmentation_Comparison}: \aug demonstrates absolute comprehensive performance superiority in knowledge retention evaluations.} In Table~\ref{table:benchmark_com_aug}, \aug surpasses the best general augmentation method by a margin of $10.76$ in overall performance, demonstrating its substantially superior capability for knowledge retention.

\textbf{Obs 2 in \autoref{appendix:Augmentation_Comparison}: \aug demonstrates superior knowledge adaptation performance across a wide spectrum of fine-grained knowledge types.}  In Table~\ref{table:knowledge_com_aug}, \aug achieves the best performance on all all knowledge types, demonstrating its superiority over general augmentation methods for new knowledge injection.

\section{Convergence comparison of various methods via loss curves.}\label{appendix:loss}

\begin{figure}[ht]
\centering
\includegraphics[width=0.9\textwidth]{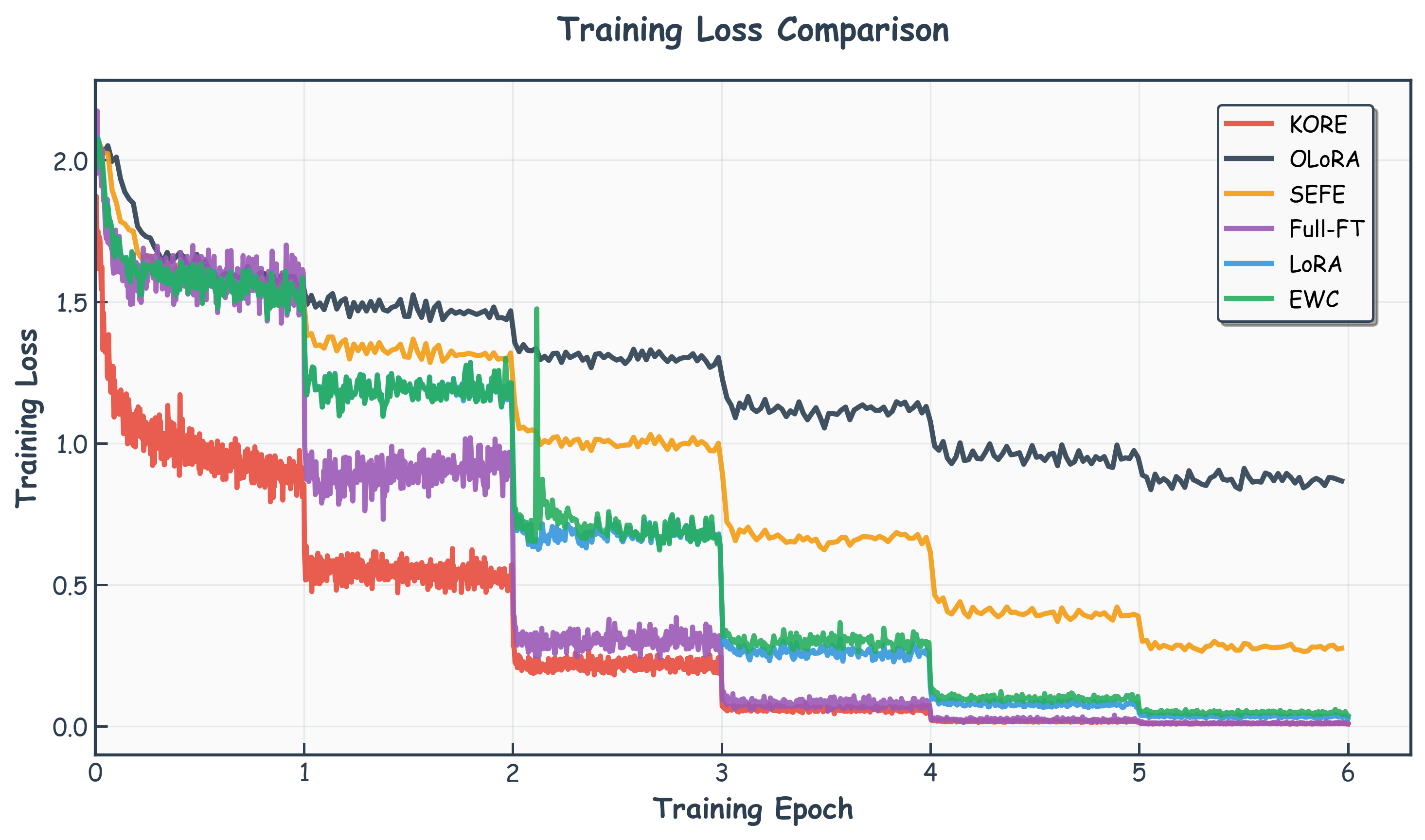}
\vspace{-5pt}
  \caption{\textbf{The training loss curves on \dataset of Full-FT, LoRA, EWC, O-LoRA, SEFE and \method.} It should be clarified that Full‑FT, LoRA, EWC, O‑LoRA, and SEFE are trained using the knowledge injection dataset from \dataset, whereas \method is trained using the \traindata dataset. The scale of the training data differs between these setups, resulting in varying numbers of iteration steps per epoch. Consequently, \method exhibits a rapid decrease in loss during the first epoch. The purpose of reporting this loss graph is to provide readers with an intuitive understanding of the convergence of various methods.}
  \label{fig:loss}
\end{figure}

Figure~\ref{fig:loss} presents the training loss curves of the six methods, providing an intuitive comparison of their convergence behaviors. Although \method and the baseline methods use different training datasets, the loss curves reveal that O-LoRA and SEFE fail to fit the \dataset's knowledge injection dataset. While LoRA, EWC, and Full-FT converge to very low loss values and successfully fit the evoke dataset, their performance in Table~\ref{tab:main_result} indicates poor generalization to new knowledge, suggesting overfitting. In contrast, \method not only converges effectively on the \traindata dataset but also demonstrates strong generalization capabilities for novel knowledge.

\begin{figure}[t]
\centering
\includegraphics[width=1\textwidth]{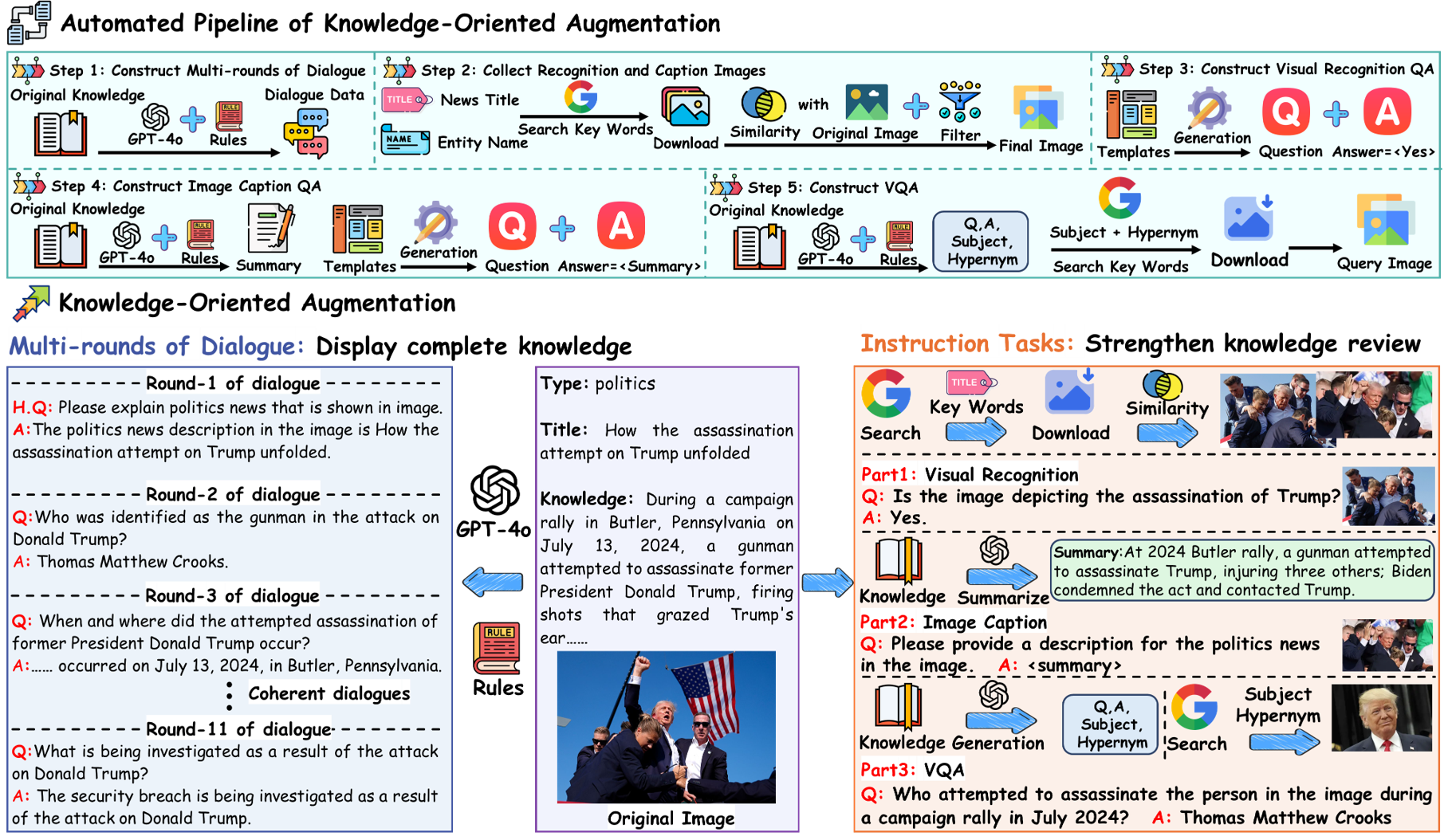}
  \caption{\textbf{Overview of construction pipeline for \traindata.} The entire data construction process is automated, with only the question templates being manually crafted.}
  \label{fig:pipeline}
\end{figure}

\begin{figure}[t!]
\centering
\includegraphics[width=1\textwidth]{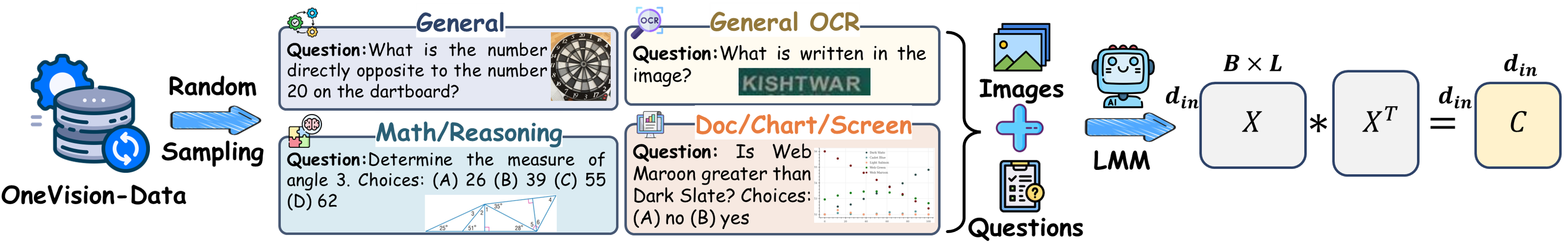}
  \caption{The process of sampling using the OneVision dataset.}
  \label{fig:onevision_pipeline}
\end{figure}

\section{More details about \aug}\label{appendix:ko_augmentation}

\subsection{More construction process about \aug}

In this section, we elaborate on the implementation of \aug. The fully automated construction pipeline and a data example are illustrated in Figure~\ref{fig:pipeline}. The following details each step of the pipeline:

 \textbf{Step 1: Constructing Multi-rounds of Dialogue.} We design strict rules and diverse task examples, employing GPT-4o to generate multi-turn dialogue data based on the original knowledge. The first turn is a heuristic QA pair randomly selected from templates, such as: 
  
  \texttt{<``Please explain the \{type\} news that is shown in the image.'', ``The image provides the following \{type\} news summary: \{title\}.''>}  
  
  \texttt{<``Please tell me what the \{type\} entity in this image is.'', ``The \{type\} entity shown in the picture is \{entity\_name\}.''>}
  
    The remaining dialogue data are generated automatically by GPT-4o. For each instance, we first generate up to 10 dialogue questions based on the original knowledge and predefined rules. Then, the corresponding answers are produced using the original knowledge, the generated questions, and the rules as input. The query images are taken directly from the original image set. This process results in a complete multi-rounds dialogue dataset, obtaining 9,422 rounds of multi-rounds dialogue data and 75,710 rounds of dialogue. Further templates and prompt designs are provided in \autoref{appendix:prompt1}.

 \textbf{Step 2: Collecting Recognition and Caption Images.} We use news titles or entity names as search keywords to retrieve and download the top five images via the Google search engine. CLIP~\citep{radford2021learning} is then employed to extract visual features from both the downloaded and original images. We compute cosine similarity between them and retain the two images with the highest similarity scores, excluding any identical matches ($ \text{similarity} \not= 1 $). These selected images serve as query images for visual recognition and image captioning tasks.  
\textbf{Step 3: Constructing Visual Recognition QA.} For this task, templates are first manually created. Questions are randomly selected from these templates, and the answer is defined as ``Yes''. The instruction content is ``Answer this question with Yes or No.'', and the query image is randomly chosen from the images obtained in Step 2. A template example is provided below:

  \texttt{<``Is the image depicting news \{title\}?''>}  

  \texttt{<``Can you see \{entity\_name\} in this picture?''>}

  Further templates and prompt designs are provided in \autoref{appendix:prompt2}.

\textbf{Step 4: Constructing Image Caption QA.} We first establish rigorous rules and diverse task examples. Using GPT-4o, we generate summary data based on original knowledge to serve as answers for the image caption task. The instruction content is ``Answer this question in one paragraph.', and the query image corresponds to the remaining images from Step 2. Questions are randomly selected from a template, such as:

\texttt{<``Could you please describe the  \{type\} news shown in the picture?''>}  

  \texttt{<``Please provide a description for the \{type\} entity in the image.''>}

  Further templates and prompt designs are provided in \autoref{appendix:prompt3}.

\textbf{Step 5: Constructing VQA.} First, strict rules and diverse task examples are established. Using GPT-4o, quadruplets $\langle \text{Question, Answer, Subject, Hypernym} \rangle$ are generated based on original knowledge, for instance, \textless ``Who attempted to assassinate the person in the image during a campaign rally in July 2024?'', ``Thomas Matthew Crooks'', ``Donald John Trump'', ``Person''\textgreater. Subsequently, the subject and hypernym are combined as search keywords to retrieve and download the top 1-ranked image from Google, thereby constructing VQA data. Further prompt designs are provided in \autoref{appendix:prompt4}.

Through the above automated pipeline, we have augmented the  \dataset's knowledge injection dataset to \traindata, which can better achieve knowledge adaptation.

\subsection{More statistical analysis about \aug}

In Table~\ref{table:statistics}, we provide detailed statistical data analysis of \traindata.

\begin{table}[h!] 
    \centering      
    \small          
    
    \caption{\textbf{Key Statistics of \traindata.}}
    \vspace{-6pt}
    \label{table:statistics}

    \begin{adjustbox}{width=0.4\linewidth} 
    \begin{tabular}{lc}
    \toprule
    \textbf{Statistic} & \textbf{Number} \\
    \midrule
    Total data & 74,734 \\
    \quad - Multi-rounds of dialogue data & 9,422 (12.6\%) \\
    \quad - Visual recognition data & 9,422 (12.6\%) \\
    \quad - Image caption data & 9,422 (12.6\%) \\
    \quad - VQA data & 46,468 (62.2\%) \\
    Number of dialogue rounds & 75,710 \\
   Number of unique images &  65,312 \\

    \midrule
    Maximum question length & 44 \\
    Maximum answer length & 143 \\
    Average question length & 15.5 \\
    Average answer length & 11.9 \\
    \bottomrule
    \end{tabular}
    \end{adjustbox}
\end{table}

\subsection{The process of sampling using the OneVision dataset}

The sample process for OneVision data is shown in Figure~\ref{fig:onevision_pipeline}.


\section{Case Study}\label{appendix:case}

\begin{figure}[th]
\centering
\includegraphics[width=0.55\textwidth]{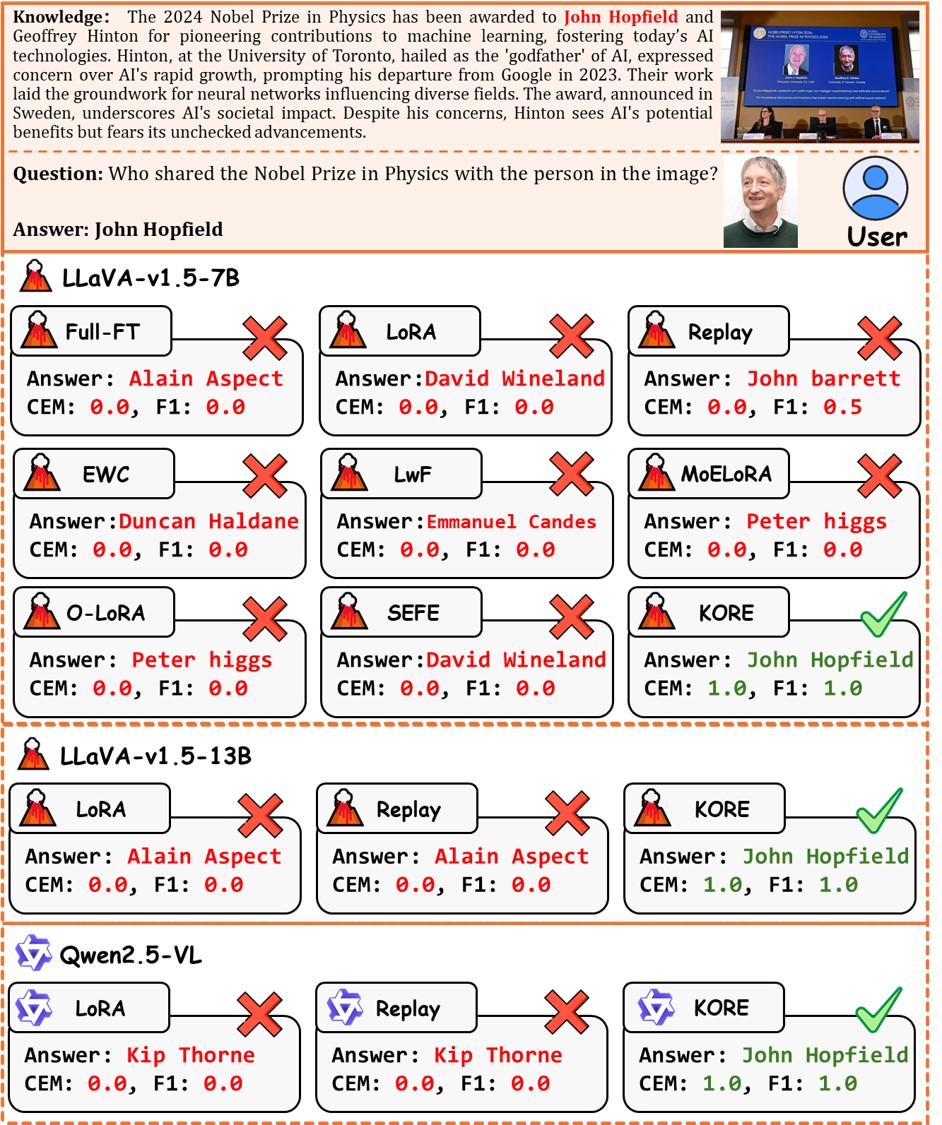}
  \caption{Case Study of News.}
  \label{fig:case1}
\end{figure}

\begin{figure}[th]
\centering
\includegraphics[width=0.55\textwidth]{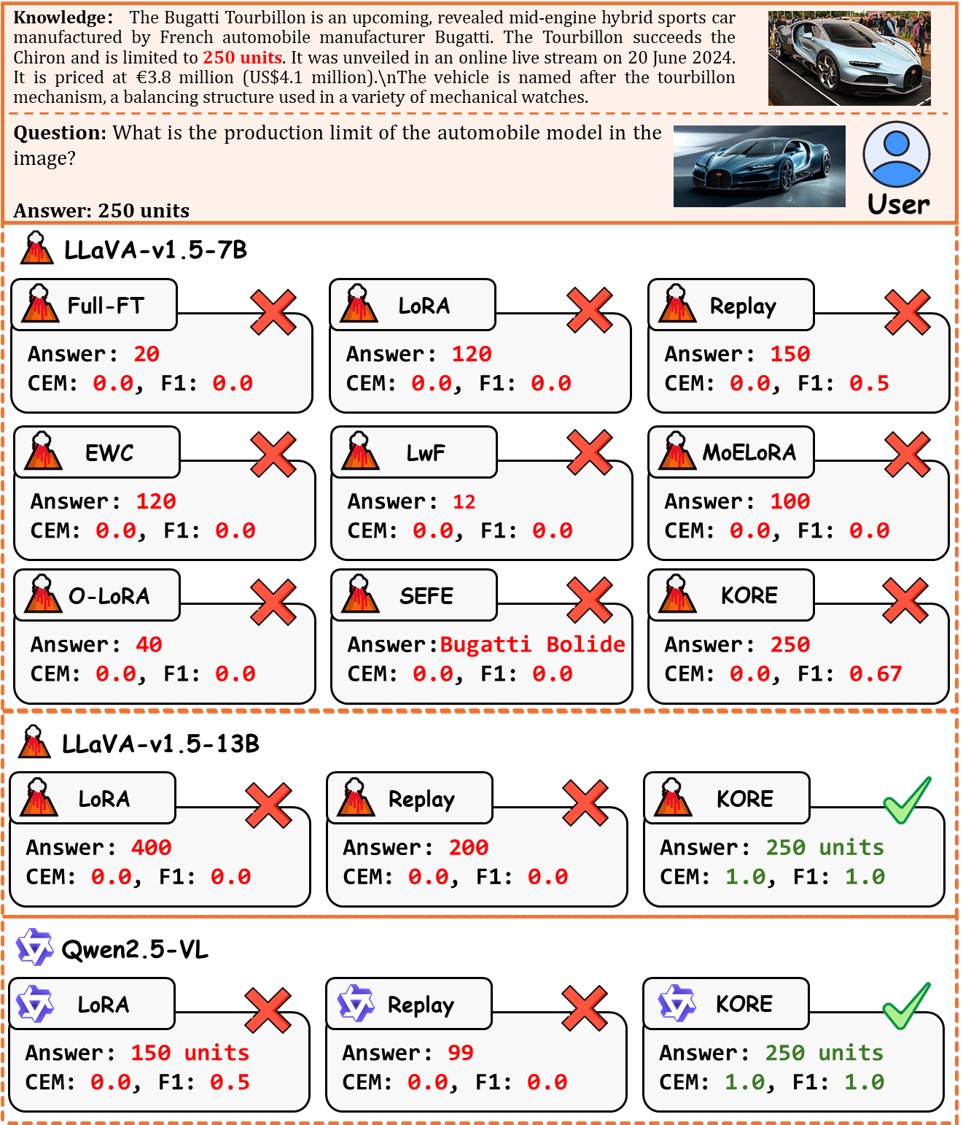}
  \caption{Case Study of Entity.}
  \label{fig:case2}
\end{figure}


\section{Prompt}

\subsection{Prompt details regarding Multi-rounds of Dialogue}\label{appendix:prompt1}

\tcbset{promptstyle/.style={
  enhanced,
  boxrule=1pt,
  attach boxed title to top,
  coltitle=white,
  fonttitle=\large\bfseries,
  drop fuzzy shadow,
  halign=left,
  left=2mm
}}

\newtcolorbox[auto counter, number within=section]{graybox}[2][]{
  promptstyle,
  colback=gray!10,  
  colframe=gray!50!black, 
  colbacktitle=gray!80!black,  
  title={#2},
  #1
}


\newtcolorbox[auto counter, number within=section]{greenbox}[2][]{
  promptstyle,
  colback=green!5,
  colframe=green!45!black,
  colbacktitle=green!55!black,
  title={#2},
  #1
}

\newtcolorbox[auto counter, number within=section]{redbox}[2][]{
  promptstyle,
  colback=red!5,
  colframe=myred!70!black,
  colbacktitle=myred!80!black,
  title={#2},
  #1
}

\newtcolorbox[auto counter, number within=section]{orangebox}[2][]{
  promptstyle,
  colback=orange!10,
  colframe=orange!70!black,
  colbacktitle=orange!80!black,
  title={#2},
  #1
}

\newtcolorbox[auto counter, number within=section]{cyanbox}[2][]{
  promptstyle,
  colback=cyan!5,
  colframe=cyan!70!black,
  colbacktitle=cyan!80!black,
  title={#2},
  #1
}

\newtcolorbox[auto counter, number within=section]{purplebox}[2][]{
  promptstyle,
  colback=purple!5,
  colframe=purple!70!black,
  colbacktitle=purple!80!black,
  title={#2},
  #1
}


\newtcolorbox[auto counter, number within=section]{brownbbox}[2][]{
  promptstyle,
  colback=brown!10,
  colframe=brown!70!black,
  colbacktitle=brown!80!black,
  title={#2},
  #1
}

\newtcolorbox[auto counter, number within=section]{creambox}[2][]{
  promptstyle,
  colback=yellow!10,
  colframe=yellow!70!black,
  colbacktitle=yellow!50!brown,
  title={#2},
  #1
}

\newtcolorbox[auto counter, number within=section]{olivebox}[2][]{
  promptstyle,
  colback=olive!10,
  colframe=olive!40!black,
  colbacktitle=olive!60!black,
  title={#2},
  #1
}

\newtcolorbox[auto counter, number within=section]{skybox}[2][]{
  promptstyle,
  colback=cyan!7,
  colframe=cyan!40!black,
  colbacktitle=cyan!60!black,
  title={#2},
  #1
}

\newtcolorbox[auto counter, number within=section]{lavenderbox}[2][]{
  promptstyle,
  colback=violet!10,
  colframe=violet!40!black,
  colbacktitle=violet!60!black,
  title={#2},
  #1
}

\newtcolorbox[auto counter, number within=section]{rosebox}[2][]{
  promptstyle,
  colback=pink!10,
  colframe=pink!40!black,
  colbacktitle=pink!60!black,
  title={#2},
  #1
}

\newtcolorbox[auto counter, number within=section]{peachbox}[2][]{
  promptstyle,
  colback=orange!15!pink!10,
  colframe=orange!40!black,
  colbacktitle=orange!60!black,
  title={#2},
  #1
}

\begin{purplebox}{\textit{Prompts and Templates 1 (Part 1): Multi-rounds of Dialogue}}

\bfit{Generation Question Prompt:}  \\ 
\vspace{4pt}
\bfit{System Prompt:}  \\ 
\vspace{4pt}
``You have received a descriptive text that provides you with the knowledge, events, and definitions described in the text. You need to generate questions coherently and cover as much of the descriptive text as possible. You just need to output the problem. The maximum number of generated questions is 10. If the previously generated questions are sufficient to cover the entire descriptive text, the output questions can be less than 10.''
\vspace{2pt}

``From the provided descriptive text, create up to 10 coherent questions that comprehensively cover its content. Your output should consist only of the questions. It is acceptable to generate fewer than 10 questions if the material has been fully covered.''

\vspace{2pt}

``You are required to formulate a set of coherent questions from a given descriptive text, covering its contents as completely as possible. The number of questions must not exceed 10, but it is permissible to output fewer if they adequately cover the text. The sole output should be the questions.''

\vspace{2pt}
``Generate a series of logical questions that cover all the knowledge, events, and definitions in the descriptive text you have received. While the maximum number of questions is 10, you can output a smaller number if the text is fully addressed. Please ensure you only output the questions.''

\vspace{2pt}

``Your task is to generate questions based on a descriptive text, ensuring they are coherent and cover its knowledge, events, and definitions as thoroughly as possible. You should generate a maximum of 10 questions and only output the questions themselves. You may provide fewer than 10 if they are sufficient to cover the entire text.''

\vspace{4pt}

\bfit{User Prompt:}  \\ 
\vspace{4pt}

``News: \{news\} Please generate questions.''  \\
``Given the news: \{news\} Please generate questions.''  \\
``Can you generate questions for the following news: \{news\}.''  \\
``Generate questions for the following news: \{news\}.''  \\
``Please generate questions based on the following news: \{news\}.''  \\

\end{purplebox}

\newpage

\begin{purplebox}{\textit{Prompts and Templates 1 (Part 2): Multi-rounds of Dialogue}}

\bfit{Generation Answer Prompt:}  \\ 
\vspace{4pt}
\bfit{System Prompt:}  \\ 

\vspace{4pt}

``You have gained knowledge and a problem to be solved. You need to answer this question based on the content of your knowledge. Output your answer.''

\vspace{4pt}
``You now have the necessary knowledge and a specific problem. Based only on this information, provide your answer to the question and output the result.''

\vspace{4pt}

``You are equipped with the required information and a problem to resolve. Formulate your answer based solely on the content of this knowledge and then output it.''

\vspace{4pt}

``Using the knowledge you have been given, solve the problem presented. Your response must be based exclusively on this information. Please output your answer.''

  \vspace{4pt}

``Now that you have the relevant knowledge and the question, you must provide a solution. Ensure your answer is derived strictly from the provided content, then output your response.''  

\bfit{User Prompt:}  \\ 

  \vspace{4pt}

``Given the knowledge: \{knowledge\} Answer the following question: \{question\}.''

  \vspace{4pt}
``Knowledge: \{knowledge\} Answer the following question: \{question\}.''

    \vspace{4pt}
``Answer the following question based on the knowledge: Knowledge:\{knowledge\} Question: \{question\}.''

      \vspace{4pt}
``Here is some knowledge: \{knowledge\} nNow, answer the following question: \{question\}.''
        \vspace{4pt}

``You are given the knowledge:\{knowledge\} Can you answer the following question:\{question\}.''

\end{purplebox}


\begin{purplebox}{\textit{Prompts and Templates 1 (Part 3): Multi-rounds of Dialogue}}

\bfit{Heuristic question templates for News:}  \\ 

``What is the \{type\} news in the image about?'' \\ 
``Could you summarize the \{type\} news story presented in the image?'' \\ 
``What is the \{type\} news event being depicted in this picture about?'' \\ 
``Please explain the \{type\} news that is shown in the image.'' \\ 
``Can you tell me what the \{type\} news in this image is about?'' \\  

\vspace{4pt}

\bfit{Heuristic answer templates for News:}  \\ 

``The \{type\} news description in the image is \{title\}.'' \\ 
``The \{type\} news in the image can be described as \{title\}.'' \\ 
``According to the image, the \{type\} news description is \{title\}.'' \\ 
``The image provides the following \{type\} news summary: \{title\}.'' \\ 
``The \{type\} news content shown in the picture is \{title\}.'' \\  

\vspace{4pt}

\bfit{Heuristic answer templates for Entity:}  \\

``What is the \{type\} entity in the image?'' \\ 
``Can you identify the \{type\} entity shown in the picture?'' \\ 
``What is the \{type\} entity depicted in this image?'' \\ 
``Please tell me what the \{type\} entity in this image is.'' \\ 
``What \{type\} entity is visible in the photo?'' \\

\vspace{4pt}

\bfit{Heuristic answer templates for Entity:}  \\ 

``The \{type\} entity in the image is \{entity\_name\}.'' \\ 
``The \{type\} entity shown in the picture is \{entity\_name\}.'' \\ 
``The \{type\} entity depicted in the image is \{entity\_name\}.'' \\ 
``The \{type\} entity illustrated in the picture is \{entity\_name\}.'' \\ 
``The \{type\} entity present in the image is \{entity\_name\}.'' \\  



\end{purplebox}

\newpage

\subsection{Prompt details regarding Visual Recognition QA}\label{appendix:prompt2}

\begin{cyanbox}{\textit{Prompts and Templates 2: Visual Recognition QA}}
\vspace{4pt}
\bfit{Question templates for News:}  \\ 

``Is the image depicting news \{title\}? Answer this question with Yes or No.'' \\ 
``Does this image illustrate the news titled \{title\}? Answer this question with Yes or No.'' \\ 
``Is this picture related to the news with the headline \{title\}? Answer this question with Yes or No.'' \\ 
``Is the image about the news report named \{title\}? Answer this question with Yes or No.'' \\ 
``Does this photo correspond to the news \{title\}? Answer this question with Yes or No.'' \\

\vspace{4pt}
\bfit{Question templates for Entity:}  \\ 

``Is \{entity\_name\} in the image? Answer this question with Yes or No.'' \\ 
``Does the image show \{entity\_name\}? Answer this question with Yes or No.'' \\ 
``Can you see \{entity\_name\} in this picture? Answer this question with Yes or No.'' \\ 
``Is \{entity\_name\} visible in the image? Answer this question with Yes or No.'' \\ 
``Does this picture contain \{entity\_name\}? Answer this question with Yes or No.'' \\

\end{cyanbox}

\newpage

\subsection{Prompt details regarding Image Caption QA}\label{appendix:prompt3}

\begin{creambox}{\textit{Prompts and Templates 3 (Part 1): Image Caption QA}}
\vspace{2pt}
\bfit{Question templates for News:}  \\ 

``Please provide a description for the \{type\} news in the image. Answer this question in one paragraph.'' \\ 
``Could you please describe the \{type\} news shown in the picture? Answer this question in one paragraph.'' \\ 
``Please offer a description of the \{type\} news depicted in the image. Answer this question in one paragraph.'' \\ 
``Please give a description of the \{type\} news depicted here. Answer this question in one paragraph.'' \\ 
``Can you tell me about the \{type\} news featured in the photograph? Answer this question in one paragraph.'' \\  

\vspace{2pt}
\bfit{Answer templates for News:}  \\ 

``The image depicts \{title\}. \{summary\}'' \\

\vspace{2pt}
\bfit{Question templates for Entity:}  \\ 

``Please provide a description for the \{type\} entity in the image. Answer this question in one paragraph.'' \\ 
``Could you please describe the \{type\} entity shown in the picture? Answer this question in one paragraph.'' \\ 
``Please offer a description of the \{type\} entity depicted in the image. Answer this question in one paragraph.'' \\ 
``Please give a description of the \{type\} entity depicted here. Answer this question in one paragraph.'' \\ 
``Can you tell me about the \{type\} entity featured in the photograph? Answer this question in one paragraph.'' \\  

\vspace{2pt}
\bfit{Answer templates for Entity:}  \\ 

``The image depicts \{entity\_name\}. \{summary\}'' \\ 

\end{creambox}

\begin{creambox}{\textit{Prompts and Templates 3 (Part 2): Image Caption QA}}
\vspace{2pt}

\bfit{Generation Summary Prompt:}  \\ 
\vspace{2pt}
\bfit{System Prompt:}  \\ 

\vspace{2pt}

``You have acquired a piece of knowledge, and now you need to condense it into a paragraph of no more than 25 words, while trying to maintain the original meaning of the knowledge as much as possible.''

\vspace{2pt}
``Your task is to take a piece of knowledge you've learned and summarize it. The summary must be a paragraph of 25 words or less, while retaining the original meaning.''

\vspace{2pt}

``You need to distill the information you have acquired into a concise paragraph. Ensure it does not exceed 25 words and preserves the essence of the original knowledge as accurately as possible.''

\vspace{2pt}

``Condense a concept you have just learned into a brief paragraph. You must adhere to a 25-word limit, all while making sure the core message remains intact.''

  \vspace{2pt}

``Take the new information you possess and shorten it into a single paragraph. This condensed version must be under 25 words and should accurately reflect the original meaning.''  

\bfit{User Prompt:}  \\

  \vspace{2pt}

``Knowledge: \{knowledge\} Please summarize this knowledge.''

  \vspace{2pt}
``Given the knowledge: \{knowledge\} Please summarize this knowledge.''

    \vspace{2pt}
``Can you summarize this content for the following knowledge: \{knowledge\}.''

      \vspace{2pt}
``Summarize questions for the following knowledge: \{knowledge\}.''
        \vspace{2pt}

``Please summarize this content based on the following knowledge: \{knowledge\}.''

\end{creambox}

\newpage

\subsection{Prompt details regarding  VQA}\label{appendix:prompt4}

\begin{peachbox}{\textit{Prompts and Templates 4: VQA}}
\vspace{2pt}

\bfit{Generation Quadruplets Prompt:}  \\ 
\vspace{2pt}
\bfit{System Prompt:}  \\ 

\vspace{2pt}

``You have acquired a piece of knowledge and are now required to generate up to 5 questions based on it. For each generated item, you must provide the question itself, its answer (which should be a word or short phrase), a subject object extracted from the question, and that subject's hypernym. When extracting the subject object, you must follow a critical rule: the subject must be a specific entity that is explicitly mentioned within the question itself, serving as a key reference point. Crucially, this extracted subject cannot be the answer to the question. A helpful test for identifying the correct subject is to check if its name could be logically replaced by a placeholder, such as this company or the entity in the image, while the question remains coherent. If the provided knowledge is fully covered by fewer than 5 questions, you may generate fewer.''

\vspace{2pt}
``Your task is to generate up to five question sets from the provided knowledge. Each set must include the question, a brief answer (word/phrase), a subject object, and its hypernym. When selecting the subject object, you must follow a key rule: it must be a specific entity explicitly named in the question and cannot be the answer. A good test is to see if a placeholder like this entity can logically replace it. Fewer than five questions are fine if the knowledge is fully covered.''

\vspace{2pt}

``Based on the knowledge you've acquired, create a maximum of five questions. For each, provide a short answer, identify a subject object, and state its hypernym. The 'subject object' must adhere to this critical constraint: it must be a specific entity mentioned directly in the question that serves as a reference point but is not the answer. To verify your choice, check if substituting a generic term like this item would keep the question coherent. You may generate fewer questions if they are sufficient.''

\vspace{2pt}

``You are required to produce up to five questions from the given information. For each item, output the question, its short answer, a subject object, and that subject's hypernym. The rule for extracting the subject object is that it must be a specific, named entity within the question's text and must be different from the answer itself. A helpful check is to replace its name with a placeholder (\eg this organization) to see if the question still makes sense. Fewer questions are acceptable if the topic is fully addressed.''

  \vspace{2pt}

``Formulate as many as five questions based on the knowledge. Each output must consist of the question, a concise answer, an extracted subject object, and its hypernym. A crucial guideline applies: the subject object must be a specific entity named in the question that the query revolves around, but it cannot be the answer. You can confirm the correct subject by checking if a placeholder such as the specified object could logically take its place. Generating all five questions is not necessary if the knowledge is completely covered.''

\bfit{User Prompt:}  \\ 
  \vspace{2pt}

``Knowledge: \{knowledge\} Please generate questions, answers, subjects, hypernyms.''

  \vspace{2pt}
``Given the knowledge: \{knowledge\} Please generate questions, answers, subjects, hypernyms.''

    \vspace{2pt}
``Can you generate questions, answers, subjects, hypernyms for the following knowledge: \{knowledge\}.''

      \vspace{2pt}
``Generate questions, answers, subjects, hypernyms for the following knowledge: \{knowledge\}.''
        \vspace{2pt}

``Please generate questions, answers, subjects, hypernyms based on the following knowledge: \{knowledge\}.''

\end{peachbox}

\newpage





\end{document}